\documentclass[10pt,twocolumn,letterpaper]{article}

\usepackage[pagenumbers]{cvpr}

\newcommand{\fycolor}{}
\usepackage{pgfplots}
\pgfplotsset{compat=1.18}
\usetikzlibrary{positioning,calc,fit,backgrounds,arrows.meta,shapes.geometric}
\usepackage{fontawesome5}
\usepackage{afterpage}
\usepackage{refcount}

\definecolor{cvprblue}{rgb}{0.21,0.49,0.74}
\usepackage[pagebackref,breaklinks,colorlinks,allcolors=cvprblue]{hyperref}
\usepackage{tabularray}
\UseTblrLibrary{booktabs}

\title{Perception Modules Meet End-to-End Driving}

\author{
Fuyang Zhang$^{1*}$\quad
Artur Parkhimchyk$^{1*}$\quad
Mehdi Esmaeilzadeh$^{1*}$\\
Eddie Sheung$^{1}$\quad
Zhengqing Wang$^{1}$\quad
Yasutaka Furukawa$^{1,2}$\\[0.4em]
$^{1}$Simon Fraser University\qquad $^{2}$Wayve\\[0.2em]
{\normalsize $^{*}$Equal contribution}
}

\begin{document}
\maketitle
\begin{abstract}
This paper presents a method that bridges modular and end-to-end autonomous driving by using perception modules as training-time teachers to compress visual backbone tokens into compact, driving-focused representations, improving end-to-end driving models without requiring perception at inference.
Concretely, we introduce two techniques: 1) \emph{semantic-segmentation guidance}, which uses a gated autoencoder to remove driving-irrelevant background components from image patch tokens; and 2) \emph{object-detection guidance}, which pulls together scene-level register tokens across driving scenes that share similar driving conditions in terms of detected objects and planned trajectories.
Our methodology is agnostic to the driving model backbone, and we apply our techniques to six end-to-end driving models across ViT and ResNet-34 architectures.
Extensive experiments on the NAVSIM benchmarks show that our method consistently improves all six models (+1.4 EPDMS on DrivoR and +1.0 on DiffusionDrive), while keeping inference overhead minimal.

\end{abstract}

{\fycolor
\section{Introduction}
\label{sec:intro}

Autonomous driving first succeeded with modular pipelines where a dedicated perception module parses sensor inputs before planning, whereas recent end-to-end models \cite{hu2023planning,chitta2023transfuser,liao2025diffusiondrive,kirby2026driving} map camera images directly to driving actions.
While end-to-end driving avoids hand-crafted interfaces, perception modules such as semantic segmentation and object detection offer natural supervision for compressing visual representations. Semantic segmentation could select driving-relevant regions while discarding background textures like sky and buildings. Object detection combined with routes and waypoints could identify when visually different scenes share a similar driving condition.
Using these perception modules to compress high-dimensional visual features in end-to-end driving models will produce compact representations that improve driving robustness and generalization.

Standard end-to-end driving models instead pass all tokens from a pretrained visual encoder, such as DINO \cite{caron2021emerging,oquab2023dinov2} or JEPA \cite{assran2023ijepa,wang2026drivejepa}, directly to the driving policy.
Since these general-purpose encoders retain fine-grained background details alongside driving cues, a policy trained on limited demonstrations easily latches onto background appearance as spurious shortcuts \cite{geirhos2020shortcut}.
Naively masking out background tokens on sky, vegetation, or buildings fails to resolve this issue: self-attention has already mixed global background context into every patch token, yielding no planning gain even with segmentation masks at test time.

This paper presents a method that bridges modular and end-to-end autonomous driving, using perception modules as training-time teachers to compress visual representations and improve end-to-end driving models without requiring perception at inference.
Our approach introduces two techniques: 1) \emph{semantic-segmentation guidance} (\Cref{fig:method-overview}a), where a gated autoencoder expands each image patch token into a higher-dimensional latent dictionary and learns a gate—supervised by an offline segmentation teacher—to subtract background components such as sky, vegetation, and buildings in place while preserving driving-relevant semantics without dropping tokens; and 2) \emph{object-detection guidance} (\Cref{fig:method-overview}b), where a contrastive objective pulls together scene-level register tokens of driving scenes that share similar driving conditions, as measured in the latent space of a frozen teacher encoding detected 3D objects along the ego's path, navigation routes, and future waypoints.
Our methodology is agnostic to the driving model backbone: we apply it to six end-to-end driving models (DrivoR \cite{kirby2026driving}, Drive-JEPA \cite{wang2026drivejepa}, LTF \cite{chitta2023transfuser}, MeanFuser \cite{wang2026meanfuser}, DiffusionDrive \cite{liao2025diffusiondrive}, and PRIX \cite{wozniak2025prix}) across ViT and ResNet architectures, consistently improving driving performance on the NAVSIM benchmarks.

In summary, our contributions are threefold:

\noindent $\bullet$ \textbf{Semantic-segmentation guidance:} A gated autoencoder that suppresses driving-irrelevant background components inside each patch token without dropping tokens.

\noindent $\bullet$ \textbf{Object-detection guidance:} A contrastive alignment method that pulls together scene-level register tokens across similar driving conditions, judged by a frozen teacher of detected objects, routes, and future trajectories.

\noindent $\bullet$ \textbf{Broad architectural gains:} Consistent improvements on NAVSIM-v2 across six base models with ViT and ResNet backbones and no test-time perception overhead.

}

\section{Related Work}
\label{sec:related-work}

\subsection{End-to-End Autonomous Driving}
\label{sec:rw-e2e-driving}

End-to-end driving methods map sensor inputs directly to future trajectories.
UniAD~\cite{hu2023planning} unifies intermediate perception, prediction, and planning within a single network, whereas later works such as TransFuser~\cite{chitta2023transfuser} directly fuse sensor features for planning without explicit perception stages.
Building on this paradigm, recent studies have explored many different directions, such as multimodal trajectory generation and scoring~\cite{liao2025diffusiondrive, guo2025ipad, li2024hydramdp, li2025hydramdppp, li2025gtrs}, world-action models~\cite{zhang2025epona, li2026drivevla, xia2026drivelaw}, and vision–language–action models~\cite{hwang2025emma, renz2025simlingo, zhou2026qwen}.
Yet despite their different formulations, these approaches focus on the downstream policy or reasoning module while taking the image encoder representation as given. In contrast, we focus on the visual representation itself before it reaches the policy head, learning a compact, driving-focused representation that plugs directly into existing end-to-end models.

\subsection{Visual Foundation Models in Driving}
\label{sec:rw-vfm}

Self-supervised pretraining has produced general-purpose visual foundation models, such as MAE~\cite{he2022mae}, DINOv2~\cite{caron2021emerging, oquab2023dinov2}, and JEPA~\cite{assran2023ijepa}, that yield dense patch tokens with rich semantic features.
In end-to-end driving, recent models increasingly replace conventional CNNs with these pretrained models, either adopting DINOv2 or video JEPA as the visual backbone~\cite{feng2025rap, kirby2026driving, wang2026drivejepa, yang2025towards} or distilling foundation model priors into latent world models~\cite{zheng2025world4drive}.
By using these pretrained visual encoders, these models can leverage rich visual priors to improve sample efficiency and generalize better to unseen driving scenes.
Despite these benefits, general-purpose encoders retain visual details across the entire image, including background regions such as sky, vegetation, and buildings, which can lead the policy head to rely on background shortcuts~\cite{geirhos2020shortcut}.
\subsection{Token and Feature Compression}
\label{sec:rw-compression-distillation}

Compressing visual tokens is a widely used strategy to reduce redundancy and computation in vision systems, pioneered in vision transformers by dropping uninformative patches as in DynamicViT~\cite{rao2021dynamicvit} or merging similar tokens as in EViT~\cite{liang2022evit} and ToMe~\cite{bolya2023tome}.
Such token reduction techniques have been widely applied across domains, including pooling or pruning visual tokens in vision-language models~\cite{xing2024pyramiddrop, chen2024image, shang2025llava}, condensing multi-view tokens in novel view synthesis~\cite{asim2026scenetok, wang2025zpressor}, and merging redundant multi-view tokens in feed-forward 3D reconstruction~\cite{chen2026co, shu2026litevggt, wang2026httm}.
Similarly, end-to-end driving methods compress visual features by replacing dense BEV grids with sparse instance queries~\cite{sun2024sparsedrive} or summarizing camera patches into a small set of register tokens~\cite{kirby2026driving, li2026ntr}.
Whereas these methods reduce token count or dimensionality to save compute, we preserve the token interface and instead compress the information content of the representation in place.
Inspired by sparse autoencoders~\cite{li2026visae} that decompose the features of a frozen model with top-$k$ sparsity for interpretability, our gated autoencoder is trained jointly with the base model.
Instead of enforcing per-token code sparsity, it uses semantic-segmentation guidance to learn a global gate that subtracts driving-irrelevant latents in place.
Object-detection guidance complements this by compressing cross-scene variation in the register tokens, pulling together scenes that share similar driving conditions.

{\fycolor
\section{Method}
\label{sec:method}
\label{sec:method-overview}

Our method compresses the visual representations of end-to-end driving models using perception modules as training-time teachers.
The approach is architecture-agnostic and applies to any base driving model, including ViT models (DrivoR~\cite{kirby2026driving}, Drive-JEPA~\cite{wang2026drivejepa}) and ResNet-34 models (LTF~\cite{chitta2023transfuser}, MeanFuser~\cite{wang2026meanfuser}, DiffusionDrive~\cite{liao2025diffusiondrive}, PRIX~\cite{wozniak2025prix}).
Each base model consists of an image encoder and a policy head (\Cref{fig:method-overview}, top).
Preserving token count and dimensionality while keeping the policy head unchanged, we compress the information content of the encoder's tokens via \emph{semantic-segmentation guidance} on patch tokens (\Cref{fig:method-overview}a, \Cref{sec:seg-teacher}) and \emph{object-detection guidance} on register tokens (\Cref{fig:method-overview}b, \Cref{sec:contrastive}), trained jointly with the driving objective (\Cref{sec:training-objective}).

\begin{figure*}[t]
\centering
\includegraphics[width=\textwidth]{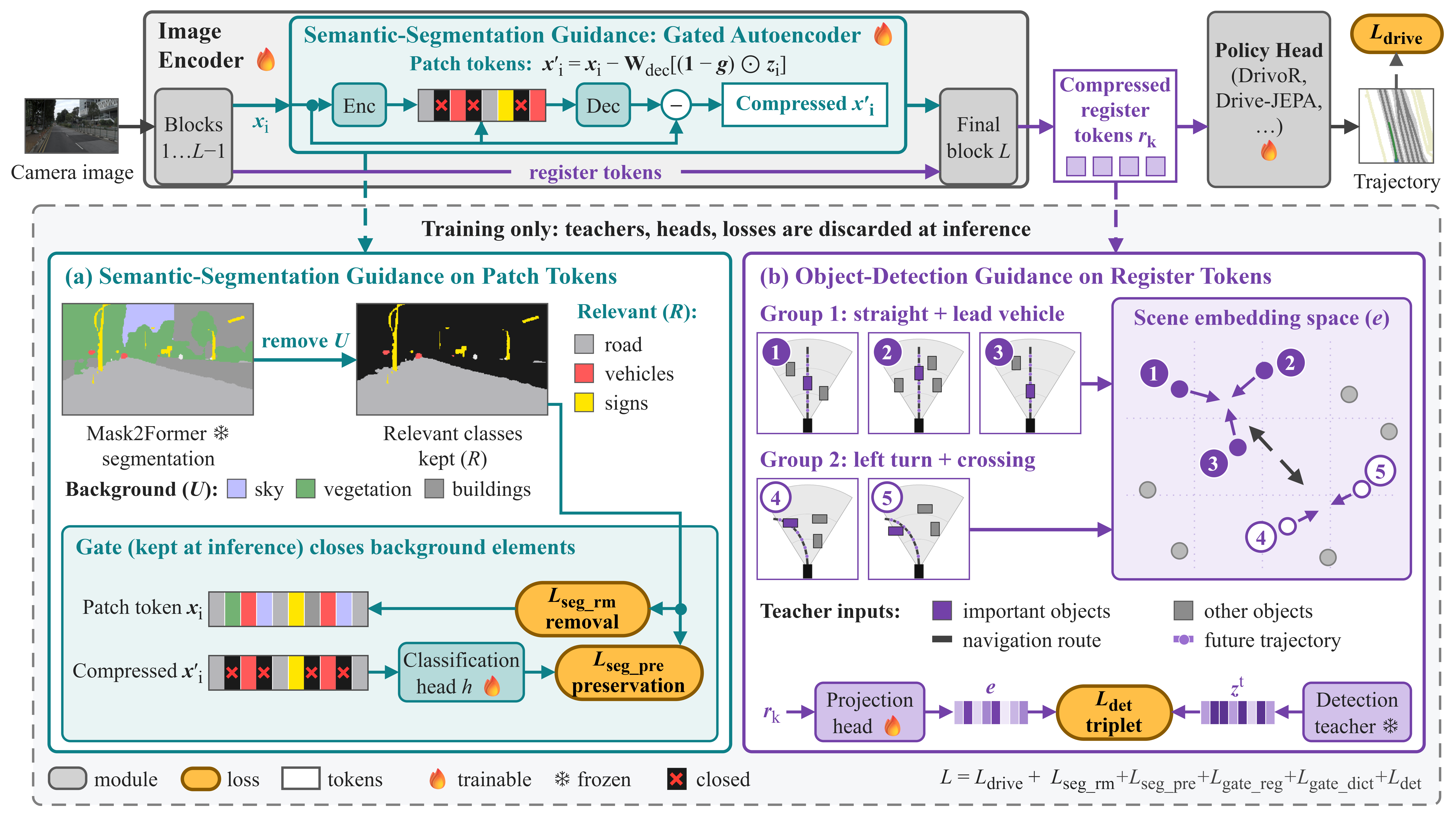}
\caption{Overview of our perception-guided representation compression method.
\textbf{Top (inference path, shown on a ViT encoder):} A gated autoencoder inserted into the image encoder compresses each patch token $\mathbf{x}_i$ into $\mathbf{x}'_i$ in place by subtracting latent channels closed by the gate $\mathbf{g}$, before the policy head reads the output tokens.
\textbf{(a)~Semantic-segmentation guidance (patch tokens):} Offline segmentation separates driving-relevant classes ($\mathcal{R}$) from background ($\mathcal{U}$, in black) to close background channels ($\mathcal{L}_{\mathrm{seg\_rm}}$) while preserving relevant semantics ($\mathcal{L}_{\mathrm{seg\_pre}}$).
\textbf{(b)~Object-detection guidance (register tokens):} A frozen teacher encodes detected objects, routes, and future trajectories into $\mathbf{z}^{\mathrm{t}}$, and a triplet loss ($\mathcal{L}_{\mathrm{det}}$) pulls together register embeddings $\mathbf{e}$ of scenes with similar driving conditions (scenes 1--3 and 4--5) while pushing distinct conditions apart.
All teachers, heads, and losses are discarded at inference.}
\label{fig:method-overview}
\label{fig:method-overview-v2}
\label{fig:teaser}
\label{fig:teaser-v2}
\end{figure*}

\subsection{Semantic-Segmentation Guidance}
\label{sec:dictionary}
\label{sec:gating}
\label{sec:seg-teacher}
\label{sec:task-supervision}

Visual details such as the color and texture of the sky or buildings do not matter for driving.
Semantic-segmentation guidance compresses each patch token in place onto a driving-relevant subspace by removing these irrelevant details while keeping driving-relevant features (\Cref{fig:method-overview}a).
Specifically, a \emph{gated autoencoder} expands each patch token into a wider latent space with a learnable gate to separate relevant and irrelevant features across channels. A \emph{segmentation teacher} optimizes this gate using patch-level semantic labels via a \emph{removal loss} and a \emph{preservation loss}.

\noindent \textbf{Gated autoencoder.} We insert the gated autoencoder before the final block on ViTs (so the final block passes compressed patch tokens into the register tokens) or after the final stage on ResNets (\Cref{fig:method-overview}, top).
Inspired by sparse autoencoders~\cite{li2026visae}, a linear autoencoder maps each patch token $\mathbf{x}_i\in\mathbb{R}^{d}$ into $m>d$ latent codes $\mathbf{z}_i\in\mathbb{R}^{m}$ and decodes them back to $\hat{\mathbf{x}}_i$ via unit-norm columns of $\mathbf{W}_{\mathrm{dec}}\in\mathbb{R}^{d\times m}$, trained with the reconstruction and $\ell_1$ sparsity loss $\mathcal{L}_{\mathrm{gate\_dict}}=\|\hat{\mathbf{x}}_i-\mathbf{x}_i\|_2^2+10^{-3}\|\mathbf{z}_i\|_1$ (normalized over the batch).
Each latent $j$ has a learnable gate $g_j=\sigma(a_j)/\sigma(a^{0})$ (forming $\mathbf{g}\in[0,1]^m$ shared across patches and scenes), where $a_j\in\mathbb{R}$ is a gate logit initialized to $a^{0}$ so all gates start open ($g_j=1$) and $\sigma(\cdot)$ is the sigmoid function.
The module updates $\mathbf{x}_i$ in place by subtracting the decoded contribution of closed latents ($\mathbf{1}\in\mathbb{R}^m$ is the all-ones vector and $\odot$ is element-wise multiplication):
\begin{equation}
    \mathbf{x}'_i =
    \mathbf{x}_i -
    \mathbf{W}_{\mathrm{dec}}
    \left[(\mathbf{1}-\mathbf{g})\odot\mathbf{z}_i\right].
    \label{eq:subtractive-gate}
\end{equation}
During training, we regularize the gate logits with a gate regularization loss $\mathcal{L}_{\mathrm{gate\_reg}}=\big(\frac{1}{m}\sum_{j=1}^{m}\sigma(a_j)-\rho_t\big)^2+\frac{1}{m}\sum_{j=1}^{m}\sigma(a_j)\big(1-\sigma(a_j)\big)$, where the first term drives the fraction of open gates toward a target ratio $\rho_t$ and the second term pushes each gate to $0$ or $1$.

\begin{figure}[t]
  \centering
  \includegraphics[width=\columnwidth]{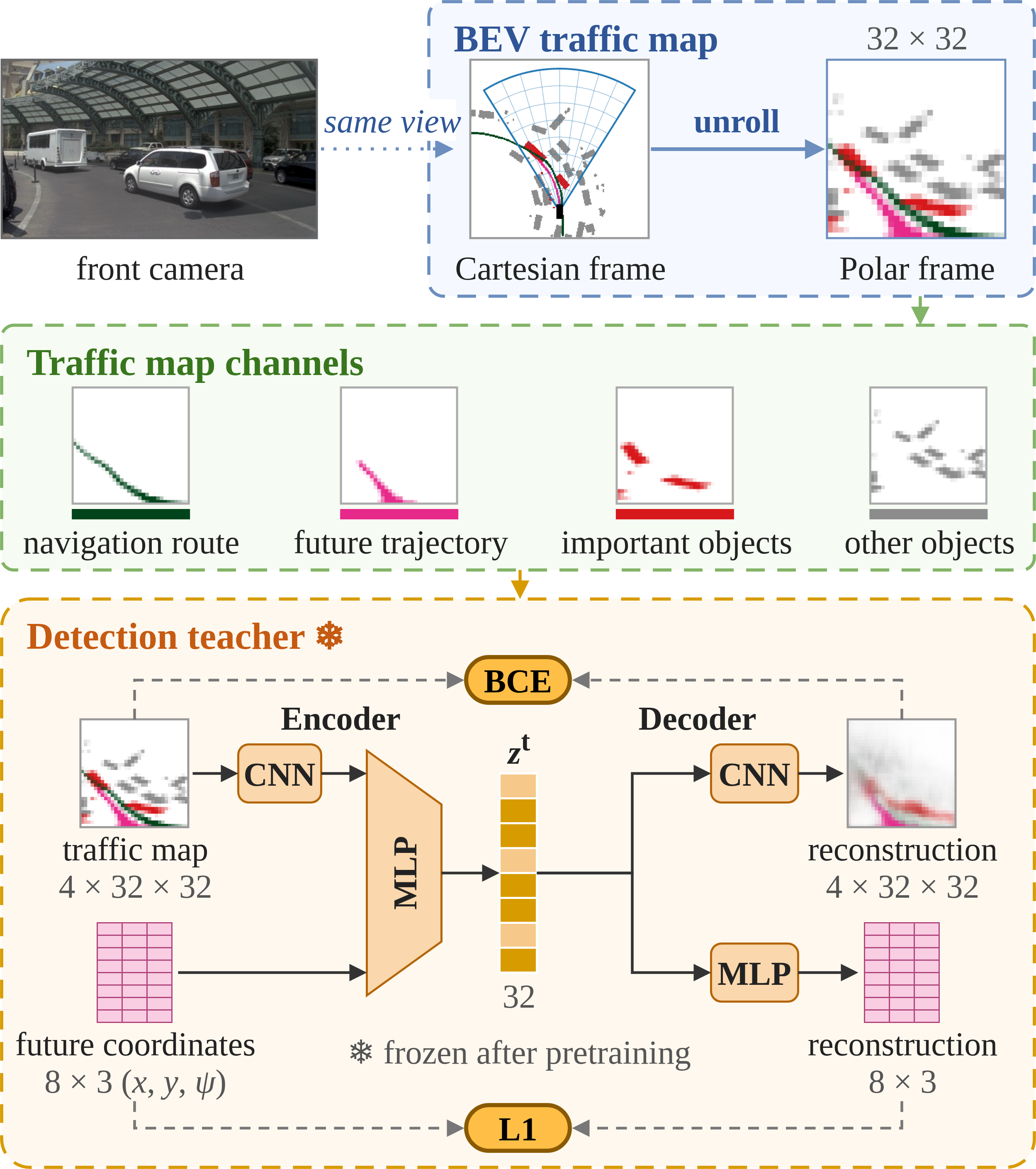}
  \caption{Object-detection guidance. Top: the BEV traffic map is unrolled from the Cartesian frame into a $32 \times 32$ polar frame only within the field of view of the camera frustum. Middle: its four traffic map channels, the teacher inputs. Bottom: the detection teacher encodes the polar BEV traffic map and the future trajectory into the latent $\mathbf{z}^{\mathrm{t}}$ and is frozen after pretraining.}
  \label{fig:map_teacher}
\end{figure}

\noindent \textbf{Segmentation teacher.}
An offline Mask2Former~\cite{cheng2022masked} teacher predicts pixel-level semantic labels and assigns each patch its majority class, grouped into driving-relevant classes $\mathcal{R}$ (road, vehicles, pedestrians, signs, and lights) and irrelevant background classes $\mathcal{U}$ (sky, vegetation, and buildings).
Using these patch labels, the teacher supervises the module through two opposite losses on the gates (\Cref{fig:method-overview}a):
1) a \emph{removal loss} ($\mathcal{L}_{\mathrm{seg\_rm}}$), which pushes to close the gates of channels linked to background classes $\mathcal{U}$ using forward-pass activation statistics; and
2) a \emph{preservation loss} ($\mathcal{L}_{\mathrm{seg\_pre}}$), which pushes to open the gates for driving-relevant classes $\mathcal{R}$ by applying a semantic segmentation cross-entropy loss to $\mathbf{x}'_i$ over only $\mathcal{R}$.

Concretely, $\mathcal{L}_{\mathrm{seg\_rm}}$ weights each gate activation $\sigma(a_j)$ by a per-channel score $s_j$:
\begin{equation}
    \mathcal{L}_{\mathrm{seg\_rm}}
    =\frac{1}{m}\sum_{j=1}^{m}s_j\,\sigma(a_j).
    \label{eq:rm-loss}
\end{equation}
From the mean forward-pass activations $\bar{z}^{\mathcal{U}}_j$ and $\bar{z}^{\mathcal{R}}_j$ of latent $j$ on patches in $\mathcal{U}$ and $\mathcal{R}$, we define $s_j$ by centering their relative difference $\tilde{s}_j$ at the median across all $m$ latents:
\begin{equation}
    \tilde{s}_j=
    \frac{\bar{z}^{\mathcal{U}}_j-\bar{z}^{\mathcal{R}}_j}
    {\bar{z}^{\mathcal{U}}_j+\bar{z}^{\mathcal{R}}_j},
    \qquad
    s_j=\tilde{s}_j-\operatorname*{median}_{k}\,\tilde{s}_k.
    \label{eq:removal-score}
\end{equation}
Latents firing more strongly on background receive $s_j>0$ and are pushed closed.
Meanwhile, $\mathcal{L}_{\mathrm{seg\_pre}}$ applies a linear classification head $h$ and cross-entropy loss $\mathrm{CE}(\cdot,\cdot)$ to $\mathbf{x}'_i$ over the relevant classes $\mathcal{C}\subseteq\mathcal{R}$ present in the batch ($I_c$ denotes patches with label $c$):
\begin{equation}
    \mathcal{L}_{\mathrm{seg\_pre}}
    =\frac{1}{|\mathcal{C}|}\sum_{c\in\mathcal{C}}
    \frac{1}{|I_c|}\sum_{i\in I_c}
    \mathrm{CE}\big(h(\mathbf{x}'_i),c\big).
    \label{eq:seg-loss}
\end{equation}

\subsection{Object-Detection Guidance}
\label{sec:contrastive}

Scenes share similar driving conditions when the ego vehicle faces similar traffic layouts along its intended path.
Object-detection guidance compresses cross-scene variation in the register tokens $\mathbf{r}_k$ by pulling such scenes into compact clusters in latent space while pushing distinct conditions apart (\Cref{fig:method-overview}b).
Our approach has three steps (\Cref{fig:map_teacher}):
1) we represent driving conditions (detected 3D boxes, navigation route, and future trajectory) as a four-channel \emph{BEV traffic map};
2) a pretrained \emph{detection teacher} autoencoder compresses each scene into a latent $\hat{\mathbf{z}}^{\mathrm{t}}$; and
3) a \emph{contrastive triplet loss} aligns the register tokens using cosine similarity of $\hat{\mathbf{z}}^{\mathrm{t}}$.

\noindent \textbf{BEV traffic map.} The BEV traffic map (\Cref{fig:map_teacher}, top and middle) has four channels: \emph{important objects} (3D detection boxes within a $3.5$\,m-wide strip along the ego's $4$\,s future trajectory), \emph{other objects} (all remaining detected boxes), the \emph{navigation route}, and the ego's \emph{future trajectory}; static road layout is omitted as it does not improve driving.
We unroll this map from the Cartesian frame into a $32 \times 32$ polar frame within the camera frustum ($\pm 32^\circ$, $2$--$48$\,m).

\noindent \textbf{Detection teacher.} The detection teacher is a lightweight autoencoder pretrained on \texttt{navtrain} for $60$ epochs and frozen (\Cref{fig:map_teacher}, bottom).
A 4-layer ConvNet encodes the $4 \times 32 \times 32$ polar traffic map into a $64 \times 2 \times 2$ feature map, which is flattened and projected to a $512$-D vector.
Since future ego motion is critical yet occupies only a few pixels in the map, we also flatten the $8$ future ego poses $(x,y,\psi)\in\mathbb{R}^{8\times 3}$ into a $24$-D vector, concatenate it with the $512$-D ConvNet vector, and pass the $536$-D vector through an MLP to obtain the unit-norm latent $\hat{\mathbf{z}}^{\mathrm{t}}\in\mathbb{R}^{32}$ (decoded by a 4-layer transposed-ConvNet and an MLP).
The cosine similarity $S^{\mathrm{t}}_{ab}=\hat{\mathbf{z}}^{\mathrm{t}\top}_a\hat{\mathbf{z}}^{\mathrm{t}}_b$ then measures driving-condition similarity.

\noindent \textbf{Contrastive triplet loss.} During driving model training, we average the register tokens $\mathbf{r}_k$, project the average to $32$ dimensions with a linear layer, and $\ell_2$-normalize it to obtain the mean register token $\mathbf{e}_a\in\mathbb{R}^{32}$ (with cosine distance $D_{ab}=1-\mathbf{e}_a^\top\mathbf{e}_b$).
For each anchor scene $a$ in a batch of size $B$, we sample pairs of scenes $(p,q)$ from the current batch and a queue of past scenes, ordering each pair by teacher similarity so $S^{\mathrm{t}}_{ap}\geq S^{\mathrm{t}}_{aq}$ (forming the set $\mathcal{P}_a$).
An adaptive-margin triplet loss~\cite{9190651} then pulls the more similar scene $p$ closer to $a$ than $q$ by a margin proportional to their teacher similarity gap $(S^{\mathrm{t}}_{ap}-S^{\mathrm{t}}_{aq})$:
\begin{equation}
    \mathcal{L}_{\mathrm{det}}
    =\frac{1}{B}\sum_{a=1}^{B}
    \frac{1}{|\mathcal{P}_a|}
    \sum_{(p,q)\in\mathcal{P}_a}
    \Big[
        D_{ap}-D_{aq}
        +\mu\big(S^{\mathrm{t}}_{ap}
        -S^{\mathrm{t}}_{aq}\big)^\gamma
    \Big]_+,
    \label{eq:contrastive}
\end{equation}
where $[\cdot]_+=\max(\cdot,0)$ and $\mu,\gamma$ control the margin scale and exponent.
This alignment acts on the native register tokens in DrivoR and on $16$ learnable register tokens appended to Drive-JEPA (fed to its policy head via a zero-initialized cross-attention layer).

\subsection{Training Objective}
\label{sec:training-objective}

Our training consists of two stages.
First, we pretrain the detection teacher autoencoder (\Cref{fig:map_teacher}, bottom) using a binary cross-entropy loss to reconstruct the $4\times 32\times 32$ polar BEV traffic map and an $\ell_1$ loss to reconstruct the $8\times 3$ future ego poses, and then freeze it.
Second, with both the segmentation teacher and detection teacher frozen, we jointly train the image encoder, gated autoencoder, and policy head with the total objective:
\begin{equation}
    \mathcal{L}
    =\mathcal{L}_{\mathrm{drive}}
    +\mathcal{L}_{\mathrm{seg\_rm}}+\mathcal{L}_{\mathrm{seg\_pre}}+\mathcal{L}_{\mathrm{gate\_reg}}+\mathcal{L}_{\mathrm{gate\_dict}}
    +\mathcal{L}_{\mathrm{det}},
    \label{eq:total-objective}
\end{equation}
where $\mathcal{L}_{\mathrm{drive}}$ is the base model's native trajectory loss (\eg, $\ell_1$ distance between predicted and ground-truth future waypoints), and the five auxiliary losses ($\mathcal{L}_{\mathrm{seg\_rm}},\mathcal{L}_{\mathrm{seg\_pre}},\mathcal{L}_{\mathrm{gate\_reg}},\mathcal{L}_{\mathrm{gate\_dict}},\mathcal{L}_{\mathrm{det}}$) are weighted by scalar coefficients $10, 1, 20, 1,$ and $1$, respectively (full formulation and teacher training details in supplementary \Cref{sec:supp-method}).
At inference, both teachers and auxiliary heads are discarded, leaving only the gated autoencoder with minimal overhead (supplementary \Cref{tab:compute}).
}

\providecommand{\figph}[3]{\fbox{\parbox[c][#2][c]{\dimexpr#1-2\fboxsep-2\fboxrule\relax}{\centering\footnotesize\color{gray}#3}}}
{\fycolor
\section{Experiments}
\label{sec:experiments}

\begin{figure*}[!t]
    \fycolor
    \centering
    \includegraphics[width=\textwidth]{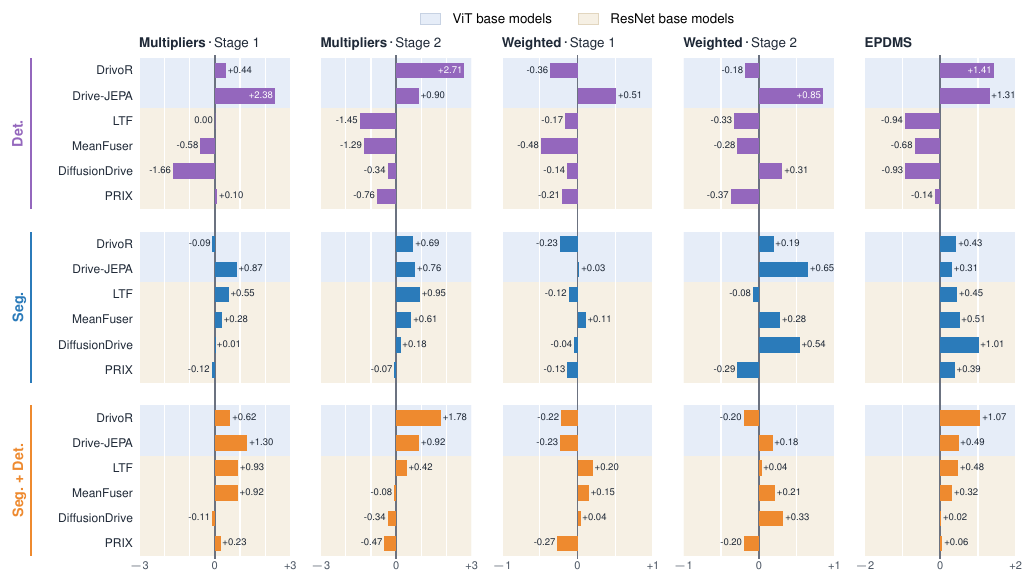}
    \caption{Effect of each guidance component. Each bar shows the difference from the base model's performance, averaged over five seeds (three for all Drive-JEPA variants and for DrivoR with Det. due to computational expenses). Multiplier metrics (Multipliers) are the product of NC, DAC, DDC, and TLC; weighted metrics (Weighted) are the weighted average of EP (5), TTC (5), LK (2), HC (2), and EC (2). Together they form the official NAVSIM EPDMS~\cite{cao2025pseudosimulation}; higher is better for all metrics. Seg.: semantic-segmentation guidance (\Cref{sec:seg-teacher}); Det.: object-detection guidance (\Cref{sec:contrastive}).}
    \label{fig:planners}
\end{figure*}

We evaluate our approach on NAVSIM-v1 and NAVSIM-v2 across six end-to-end driving base models with ViT and ResNet backbones.
We first introduce the benchmarks and experimental setup (\Cref{sec:benchmarks,sec:experimental-setup}), then present the main results (\Cref{sec:main-results}), and finally conduct ablation studies and representation analysis (\Cref{sec:ablations,sec:representation-analysis}).

\subsection{Benchmarks and Metrics}
\label{sec:benchmarks}

\noindent \textbf{NAVSIM-v1.} To verify whether nominal driving performance is preserved, we report the Predictive Driver Model Score (PDMS) on the \texttt{navtest} split of NAVSIM-v1~\cite{dauner2024navsim}.

\noindent \textbf{NAVSIM-v2.} While NAVSIM-v1~\cite{dauner2024navsim} is a standard benchmark for end-to-end driving, it evaluates policies only from nominal expert states in a single stage; as a result, its \texttt{navtest} split is largely saturated (with top methods exceeding $90$ PDMS) and does not test recovery when a policy drifts off the expert trajectory or relies on background shortcuts.
We therefore use NAVSIM-v2~\cite{cao2025pseudosimulation} on the challenging \texttt{navhard-two-stage} split as our primary benchmark.
NAVSIM-v2 evaluates policies in a two-stage closed-loop pseudo-simulation that renders novel camera views from perturbed ego poses via 3D Gaussian Splatting, directly testing out-of-distribution generalization and recovery under the Extended Predictive Driver Model Score (EPDMS).

\noindent \textbf{Reporting protocol.} Unless otherwise stated, NAVSIM-v2 results are reported at the final epoch as mean and standard deviation across five random seeds. Baselines and their corresponding variants follow identical training recipes, without test-set checkpoint selection.

\subsection{Experimental Setup}
\label{sec:experimental-setup}

\noindent \textbf{Base models.} DrivoR~\cite{kirby2026driving} is a recent end-to-end driving model with state-of-the-art performance and serves as our primary base model.
It uses a DINOv2 ViT-S/14 backbone with rank-32 LoRA and $16$ register tokens, trained on \texttt{navtrain} for $10$ epochs with AdamW, learning rate $2\times 10^{-4}$ ($10\%$ linear warm-up and cosine decay), and batch size $B=64$.
To assess generality, we integrate our method into a second ViT model, Drive-JEPA~\cite{wang2026drivejepa} ($20$ epochs, batch size $64$, $16$ added register tokens), and four ResNet-34 base models (Latent TransFuser (LTF)~\cite{chitta2023transfuser}, DiffusionDrive~\cite{liao2025diffusiondrive}, PRIX~\cite{wozniak2025prix}, and MeanFuser~\cite{wang2026meanfuser}) trained for $100$ epochs on the $8\times 32$ spatial tokens of the final ResNet stage, each following its official training recipe.

\noindent \textbf{Guidance and training.} For semantic-segmentation guidance (\Cref{sec:seg-teacher}), the gated autoencoder is placed at ViT block $10$ on DrivoR and block $22$ on Drive-JEPA (or after the final ResNet stage) with $m=4d$ latents ($m=1536$ for DrivoR with $d=384$; $m=2048$ for ResNet-34 with $d=512$), initial gate logit $a^0=2$, and gate learning rate $6\times 10^{-3}$.
We anneal $\rho_t$ from $1$ to $\rho=0.6$ between $10\%$ and $40\%$ of training and use binary STE gates in the final $25\%$ (with EMA decay $0.99$ for activation statistics).
For object-detection guidance (\Cref{sec:contrastive}), the linear projection head uses learning rate $5\times 10^{-3}$ with margin $\mu=0.4$, exponent $\gamma=1$, $64$ sampled pairs per anchor, and a FIFO queue of $512$ past scenes (full implementation details in supplementary \Cref{app:implementation-details}).

\subsection{Main Results}
\label{sec:main-results}

\xglobal\definecolor{bevRef}{RGB}{209,73,91}
\xglobal\definecolor{bevOurs}{RGB}{42,157,75}
\newlength{\bevw}
\begin{figure*}[t]
  \fycolor
  \centering
  \setlength{\bevw}{\dimexpr(\textwidth-22pt)/4\relax}
  \setlength{\tabcolsep}{0pt}
  \renewcommand{\arraystretch}{0.6}
  \newcommand{\bevimg}[1]{\includegraphics[width=\bevw]{figs/bev/#1_cam.jpg} & \includegraphics[width=\bevw]{figs/bev/#1_bev.png}}
  \newcommand{\bevlab}[2]{\multicolumn{2}{c}{\scriptsize{\color{bevRef}Baseline: #1}\quad{\color{bevOurs}+ Ours: #2}}}
  \newcommand{\bevname}[1]{\rotatebox{90}{\makebox[0.5625\bevw]{\small #1}}}
  \begin{tabular}{@{}c@{\hspace{3pt}}c@{\hspace{1pt}}c@{\hspace{6pt}}c@{\hspace{1pt}}c@{}}
    \bevname{DrivoR} & \bevimg{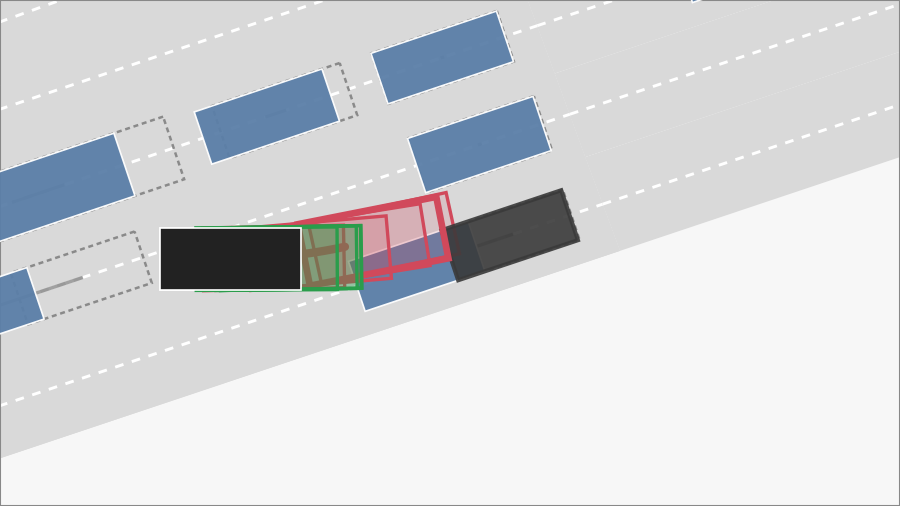} & \bevimg{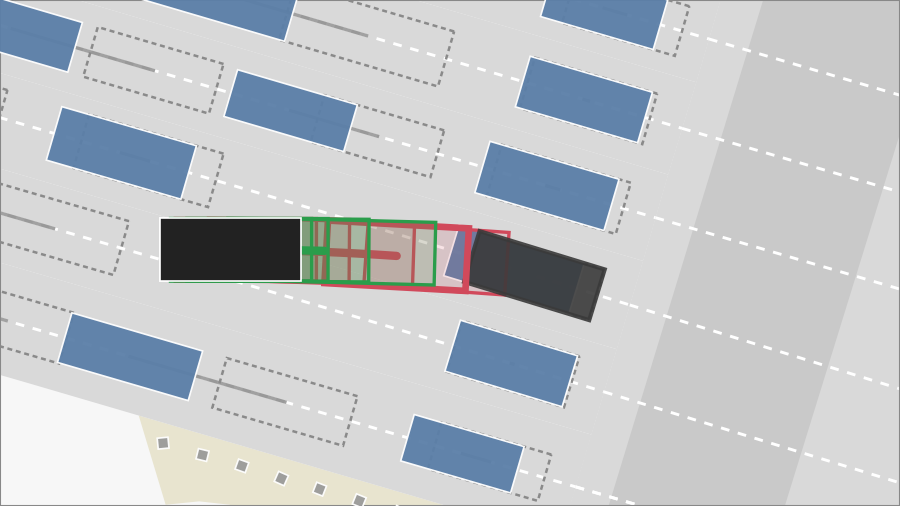} \\
    & \bevlab{EPDMS 0 (collision)}{EPDMS 100} & \bevlab{EPDMS 0 (collision)}{EPDMS 86} \\[3pt]
    \bevname{LTF} & \bevimg{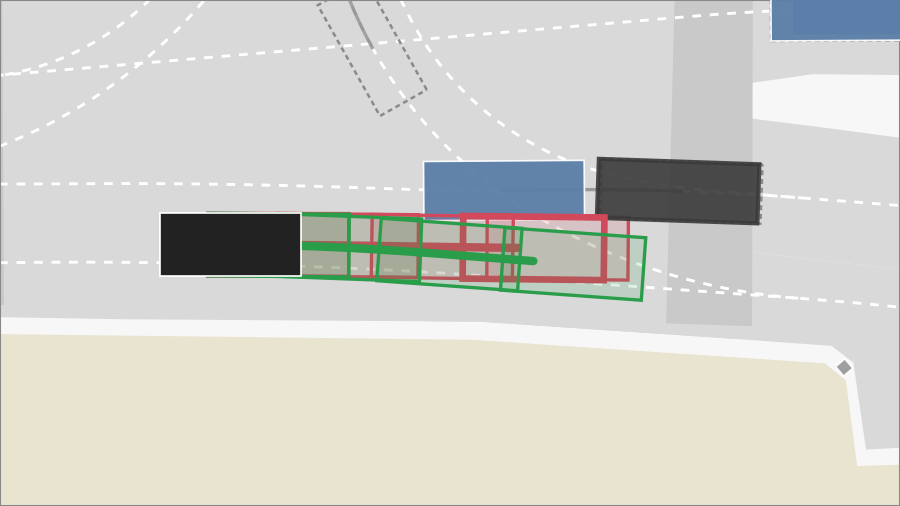} & \bevimg{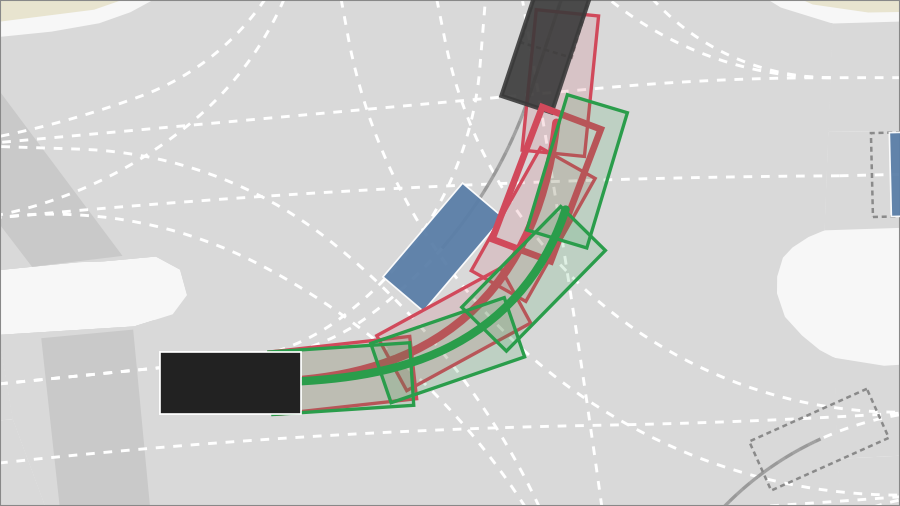} \\
    & \bevlab{EPDMS 0 (collision)}{EPDMS 96} & \bevlab{EPDMS 0 (collision)}{EPDMS 100} \\[3pt]
    \bevname{DiffusionDrive} & \bevimg{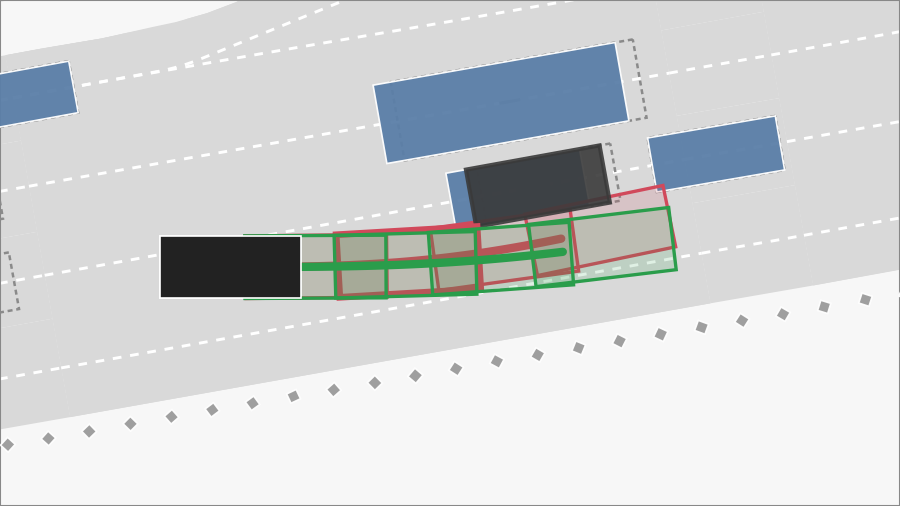} & \bevimg{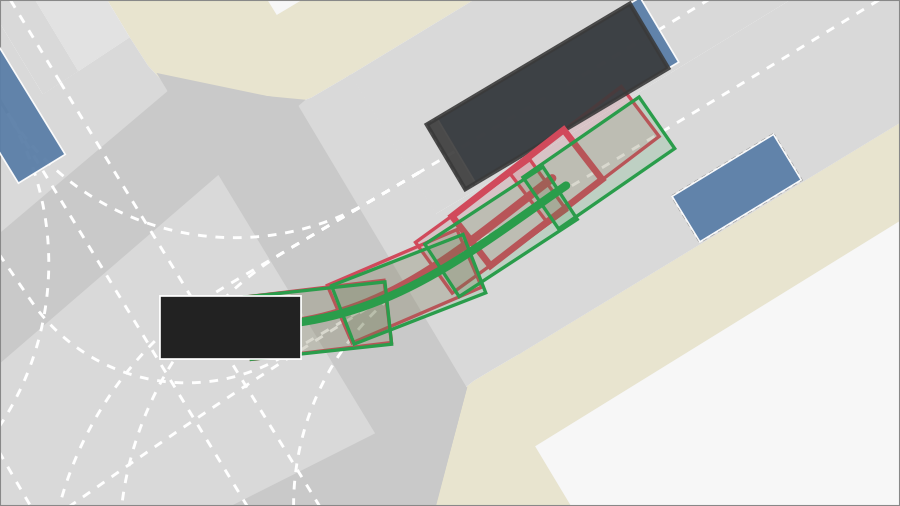} \\
    & \bevlab{EPDMS 0 (collision)}{EPDMS 100} & \bevlab{EPDMS 0 (collision)}{EPDMS 100} \\
  \end{tabular}
  \vspace{-3pt}
  \caption{Navhard scene visualization. Each example shows the front-camera rendering and BEV rollout.
  Red and green trajectories denote the baseline and our method (Seg.\ + Det.\ for DrivoR; Seg.\ for LTF and DiffusionDrive). The black rectangle marks the ego's
  current pose; colored rectangles along each trajectory show its predicted footprint at 1\,s intervals.
  Other agents are shown by their current positions (blue), trajectories (gray), and 4\,s positions
  (dashed), with the agent struck by the baseline highlighted in dark gray at contact. In each scene,
  the baseline collides in at least four of five seeds while our method avoids collision
  in at least four; one seed is shown with its per-scene EPDMS below.}
  \label{fig:bev}
\end{figure*}

\Cref{fig:planners} shows how our two guidance components change navhard
EPDMS and its stage-wise multiplier and weighted metrics across six base
models (two ViT and four ResNet-34), each relative to its own baseline trained
with the same recipe (absolute scores in supplementary \Cref{tab:supp-planners}).
The two components act on different token streams, and each helps when applied
to the tokens it is designed for.

Semantic-segmentation guidance (Seg.)\ acts on the patch tokens through the
gated autoencoder and improves EPDMS on all six base models, by $+0.31$ to
$+1.01$ points.
On DrivoR, it raises navhard EPDMS from 46.3 to 46.7.
Its gains come mainly from the Stage~2 multiplier metrics, which improve on five
of the six base models, while the weighted metrics change by less than one point.
\Cref{fig:bev} shows perturbed-start scenes on DrivoR, LTF, and
DiffusionDrive where the baseline collides in at least four of five seeds and
our model drives safely in at least four.

Object-detection guidance (Det.)\ acts on scene-level tokens.
On DrivoR, guidance on scene register tokens raises EPDMS
by $+1.41$, from 46.3 to 47.7.
Drive-JEPA has no such tokens, so we append 16 learnable register tokens to its
encoder, raising EPDMS by $+1.31$, from 32.2 to 33.5.
Combining both components (Seg.\ + Det.)\ also improves both ViT base models,
whose dedicated register tokens let the two guidance branches operate on
separate token streams ($+1.07$ on DrivoR and $+0.49$ on Drive-JEPA), but stays
below Det.\ alone.
By contrast, the four ResNet-34 base models have no register tokens, so object-detection guidance acts on their mean-pooled spatial features, which the policy head also reads.
On its own, its gradient changes the image encoder and lowers EPDMS on all four ($-0.14$ to $-0.94$).
In Seg.\ + Det., we therefore stop this gradient at the encoder features so that the detection teacher only refines the gate learned with semantic-segmentation guidance (supplementary \Cref{sec:supp-method}).
This keeps Seg.\ + Det.\ above the baselines but adds no gain over Seg.\ alone.
On NAVSIM-v1 \texttt{navtest}, every variant stays within $0.3$ PDMS of its baseline on all base models (supplementary \Cref{tab:supp-navsim-v1}).
Thus, both guidance components largely preserve nominal driving performance.
}

\begin{table}[t]
  \centering
  \footnotesize
  \renewcommand{\arraystretch}{0.90}
  \caption{NAVSIM-v1 results on \texttt{navtest}. PDMS multiplies the two multiplier metrics (NC, DAC) by the weighted average of the weighted metrics TTC (5), Comf.\ (2), and EP (5); C: camera, L: LiDAR. Rows above the last block are as reported in \cite{kirby2026driving} (four cameras). Our rows: last epoch, mean $\pm$ std over five seeds.}
  \label{tab:navsim-v1}
  \setlength{\tabcolsep}{0pt}
  \begin{tabular*}{\columnwidth}{@{\extracolsep{\fill}}llcccccc@{}}
    \toprule
    & & \multicolumn{2}{c}{Multipliers ($\times$)} & \multicolumn{3}{c}{Weighted} & \\
    \cmidrule(lr){3-4}\cmidrule(lr){5-7}
    Method & Inp. & NC$\uparrow$ & DAC$\uparrow$ & TTC$\uparrow$ & Comf.$\uparrow$ & EP$\uparrow$ & PDMS$\uparrow$ \\
    \midrule
    PDM-Closed \cite{dauner2023parting} & -- & 94.6 & 99.8 & 89.9 & 86.9 & 99.9 & 89.1 \\
    Human \cite{dauner2024navsim} & -- & 100 & 100 & 100 & 99.9 & 87.5 & 94.8 \\
    \midrule
    TransFuser \cite{chitta2023transfuser} & C+L & 97.7 & 92.8 & 92.8 & 100 & 79.2 & 84.0 \\
    DiffusionDrive \cite{liao2025diffusiondrive} & C+L & 98.2 & 96.2 & 94.7 & 100 & 82.2 & 88.1 \\
    \midrule
    Ego-status MLP \cite{dauner2024navsim} & C & 93.0 & 77.3 & 83.6 & 100 & 62.8 & 65.6 \\
    UniAD \cite{hu2023planning} & C & 97.8 & 91.9 & 92.9 & 100 & 78.8 & 83.4 \\
    LTF \cite{chitta2023transfuser} & C & 97.4 & 92.8 & 92.4 & 100 & 79.0 & 83.8 \\
    PARA-Drive \cite{weng2024paradrive} & C & 97.9 & 92.4 & 93.0 & 99.8 & 79.3 & 84.0 \\
    PRIX \cite{wozniak2025prix} & C & 98.1 & 96.3 & 94.1 & 100 & 82.3 & 87.8 \\
    Hydra-MDP++ \cite{li2025hydramdppp} & C & 98.6 & 98.6 & 95.1 & 100 & 85.7 & 91.0 \\
    iPad \cite{guo2025ipad} & C & 98.6 & 98.3 & 94.9 & 100 & 88.0 & 91.7 \\
    DriveSuprim \cite{yao2025drivesuprim} & C & 98.6 & 98.6 & 95.5 & 100 & 91.3 & 93.5 \\
    DrivoR \cite{kirby2026driving} & C & 98.9 & 98.3 & 96.2 & 100 & 89.1 & 93.1 \\
    \midrule
    DrivoR (1 cam, ours) & C & 99.1 & 98.2 & 96.9 & 100 & 84.8 & 91.2\,{\scriptsize$\pm$0.2} \\
    Ours (1 cam) & C & 99.1 & 98.3 & 96.9 & 100 & 85.3 & 91.5\,{\scriptsize$\pm$0.2} \\
    \bottomrule
  \end{tabular*}
\end{table}

\Cref{tab:navsim-v1} compares our single-camera runs with published
multi-camera results, though sensor differences preclude direct comparison.
NAVSIM-v1 is largely saturated and does not test how prediction errors
affect subsequent planning, motivating our focus on NAVSIM-v2.
Nevertheless, our single-camera approach performs well:
while DrivoR reports 93.1 PDMS, our reproduction achieves
$91.2 \pm 0.2$, which our guidance improves to $91.5 \pm 0.2$.
The gain comes mainly from ego progress (84.8 to 85.3),
with safety and comfort essentially unchanged, preserving nominal
performance alongside the NAVSIM-v2 robustness gains.

\definecolor{rtBase}{RGB}{120,120,120}
\definecolor{rtOurs}{RGB}{40,140,60}
\newcommand{\rtw}{0.134\textwidth}
\newcommand{\rtimg}[2]{{\setlength{\fboxsep}{0pt}\setlength{\fboxrule}{0.8pt}\fcolorbox{#1}{white}{\includegraphics[width=\rtw]{#2}}}}
\newcommand{\rtcol}[3]{\rtimg{#1}{figs/retrieval/#2_cam.jpg}\\[-0.6pt]\rtimg{#1}{figs/retrieval/#2_target.png}\\[-1pt]{\scriptsize\color{#1}#3}}
\newcommand{\rtcell}[3]{\begin{tabular}[t]{@{}c@{}}\rtcol{#1}{#2}{#3}\end{tabular}}
\begin{figure*}[t]
  \centering
  \setlength{\tabcolsep}{0.8pt}
  \begin{tabular}{@{}c@{\hspace{5pt}}ccc@{\hspace{5pt}}ccc@{}}
    \small Query & \multicolumn{3}{c}{\small\color{rtBase} DrivoR (baseline)} & \multicolumn{3}{c}{\small\color{rtOurs} DrivoR + Ours (Seg.\ + Det.)} \\[-1pt]
    \cmidrule(r{3pt}){1-1}\cmidrule(lr{3pt}){2-4}\cmidrule(l){5-7}
    \rtcell{black}{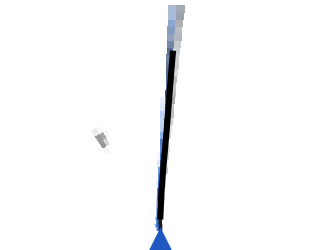}{query} &
    \rtcell{rtBase}{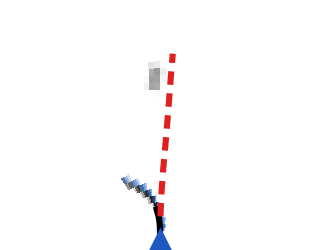}{0.44 / 19.6\,m} & \rtcell{rtBase}{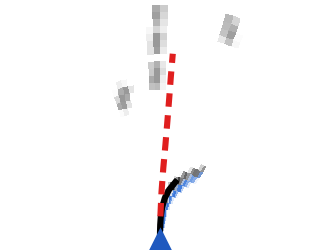}{0.38 / 15.7\,m} & \rtcell{rtBase}{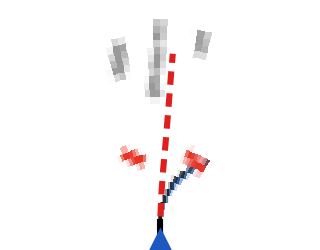}{0.31 / 20.6\,m} &
    \rtcell{rtOurs}{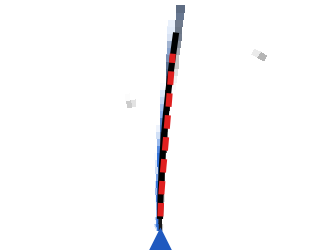}{0.97 / 1.5\,m} & \rtcell{rtOurs}{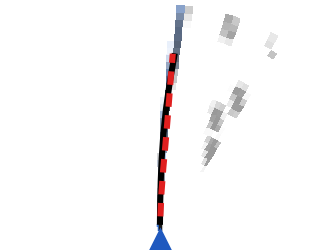}{0.90 / 2.0\,m} & \rtcell{rtOurs}{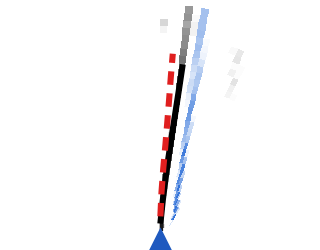}{0.92 / 2.6\,m} \\[3pt]
    \rtcell{black}{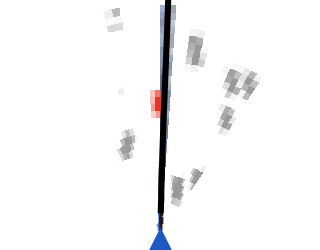}{query} &
    \rtcell{rtBase}{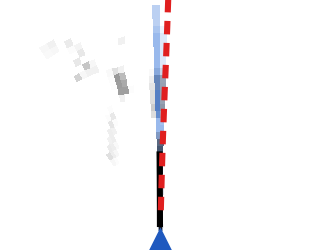}{0.39 / 16.1\,m} & \rtcell{rtBase}{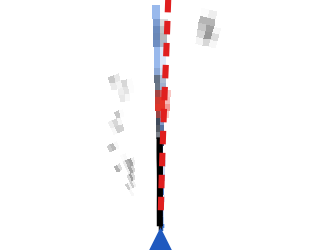}{0.34 / 14.7\,m} & \rtcell{rtBase}{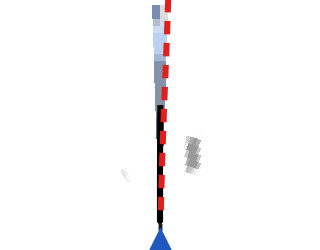}{0.54 / 11.0\,m} &
    \rtcell{rtOurs}{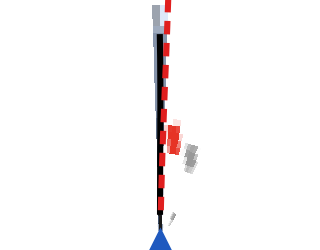}{0.92 / 3.1\,m} & \rtcell{rtOurs}{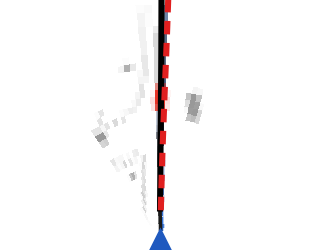}{0.98 / 1.8\,m} & \rtcell{rtOurs}{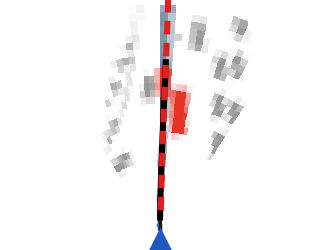}{0.89 / 6.7\,m} \\
  \end{tabular}
  \caption[Scene retrieval on navtest scenes.]{Scene retrieval on \texttt{navtest} scenes. For each query, we show the three nearest neighbors retrieved using DrivoR's register tokens before (DrivoR, gray) and after applying our method with both semantic-segmentation and object-detection guidance (DrivoR + Ours, green). Each scene is visualized by its front-camera image and BEV traffic map (detected objects, navigation route, and the scene's future trajectory in solid black, with the query's future trajectory in dashed red). Numbers report teacher cosine similarity\,$\uparrow$ / mean 4\,s waypoint distance\,$\downarrow$. Our method retrieves neighbors with closer driving conditions and future trajectories.}
  \label{fig:retrieval}
\end{figure*}

{\fycolor
\subsection{Ablation Studies}
\label{sec:ablations}

\noindent \textbf{Comparison with heuristic token-masking baselines.} A simple way to remove background information is to mask patch tokens using semantic segmentation.
In \Cref{tab:removal}, we compare our semantic-segmentation guidance on DrivoR against four token-masking baselines applied at either the ViT input or block~10. \emph{Rule} scales each patch token $\mathbf{x}_i$ by $(1 - f_i)$, where $f_i \in [0,1]$ is the fraction of background pixels (sky, vegetation, and buildings) from Mask2Former in patch $i$ (\eg, if $90\%$ of a patch belongs to sky pixels, its token is multiplied by $0.1$). \emph{Random} instead zeroes out an equivalent total number of randomly selected tokens $k = \operatorname{round}(\sum_i f_i)$.
Masking at the input substantially degrades performance ($39.7$ random and $42.9$ rule vs.\ $46.3$ for DrivoR), suggesting early self-attention blocks benefit from full-image context.
Masking at block~10 also underperforms DrivoR ($45.1$ rule and $45.3$ random), since self-attention has already mixed background and foreground features across all patch tokens.
In contrast, our semantic-segmentation guidance suppresses background channels in the expanded latent space of each token, improving over DrivoR ($46.7$ vs.\ $46.3$), consistent with the gains across all six base models in \Cref{fig:planners}.
}

\begin{table}[t]
    \fycolor
    \centering
    \caption{Comparison with heuristic token-masking baselines on
    \texttt{navhard-two-stage} using DrivoR as the base model. \emph{Rule}
    scales patch tokens by $1-f_i$ using the background pixel fraction $f_i$;
    \emph{Random} zeroes out $\operatorname{round}(\sum_i f_i)$ random tokens.}
    \label{tab:removal}
    \footnotesize
    \setlength{\tabcolsep}{0pt}
    \newcommand{\sd}[1]{\,{\scriptsize(#1)}}
    \begin{tabular*}{\columnwidth}{@{\extracolsep{\fill}}lllc@{}}
        \toprule
        Method & Masking & Where & EPDMS$\uparrow$ \\
        \midrule
        Heuristic token-masking & Random & Input & 39.7 \\
         & Random & Block 10 & 45.3 \\
         & Rule & Input & 42.9 \\
         & Rule & Block 10 & 45.1 \\
        \midrule
        DrivoR & -- & -- & 46.3\sd{1.0} \\
        DrivoR + Ours (Seg.) & -- & -- & \textbf{46.7}\sd{1.4} \\
        \bottomrule
    \end{tabular*}
\end{table}

\begin{figure}[!tb]
    \fycolor
    \centering
    \includegraphics[width=\columnwidth]{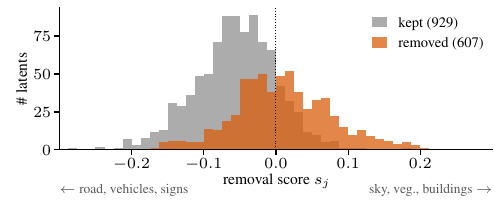}
    \caption{Latent channel selection by the gates. Distribution of the removal score $s_j$ (\Cref{eq:removal-score}) for latent channels removed and retained in DrivoR with semantic-segmentation guidance.}
    \label{fig:removal}
\end{figure}

{\fycolor
\subsection{Representation Analysis}
\label{sec:representation-analysis}

\noindent \textbf{Driving-condition retrieval on unseen scenes.} We evaluate whether scenes with similar register-token features
also share similar driving conditions on 3000 unseen
\texttt{navtest} scenes from 135 drives.
For DrivoR, we average the final-block register tokens;
for Seg.\ + Det., we additionally apply the learned 32-dimensional
projection used by $\mathcal{L}_{\mathrm{det}}$.
We compare scene similarities in these feature spaces against
those in the detection teacher's embedding space.
For retrieval, we select the five nearest neighbors, excluding
scenes from the query's driving run.

\Cref{tab:condition-retrieval} shows that Seg.\ + Det.\
increases pairwise similarity rank correlation with the teacher
from 0.351 to 0.649.
Its retrieved neighbors have higher mean teacher similarity
(0.787 to 0.812) and closer future trajectories
(4.05\,m to 3.44\,m).
These results indicate that the learned projected embedding
groups scenes by teacher-defined driving conditions.
\Cref{fig:retrieval} illustrates examples where DrivoR retrieves
visually similar scenes with different driving conditions,
while Seg.\ + Det.\ retrieves scenes with more closely
matching conditions and trajectories.
}

\begin{table}[!t]
    \fycolor
    \centering
    \caption{Driving-condition retrieval on 3000 unseen
    \texttt{navtest} scenes.
    Features: mean-pooled final-block register tokens for DrivoR;
    the learned 32-D projection of these tokens for DrivoR + Ours
    (Seg.\ + Det.).
    Rank corr.: Spearman correlation between feature-based and
    teacher-based scene-pair similarities.
    5-NN sim.: mean teacher similarity between each query and its
    five feature-retrieved neighbors.
    Traj.: mean 4\,s waypoint distance between their future ego
    trajectories and the query's.
    Retrieval excludes scenes from the query's driving run.}
    \label{tab:condition-retrieval}
    \footnotesize
    \setlength{\tabcolsep}{0pt}
    \begin{tabular*}{\columnwidth}{@{\extracolsep{\fill}}lccc@{}}
        \toprule
        Method & Rank corr.$\uparrow$ & 5-NN sim.$\uparrow$ & Traj.$\downarrow$ \\
        \midrule
        Random & -- & 0.534 & 8.31\,m \\
        DrivoR & 0.351 & 0.787 & 4.05\,m \\
        DrivoR + Ours & \textbf{0.649} & \textbf{0.812} & \textbf{3.44\,m} \\
        \bottomrule
    \end{tabular*}
\end{table}

{\fycolor
\noindent \textbf{Latent channel selection by the gates.} \Cref{fig:removal} compares the removal scores $s_j$
(\Cref{eq:removal-score}) of latent channels retained and removed by the learned gate.
Positive scores indicate stronger association with background classes, whereas negative scores reflect stronger association with driving-relevant classes.
Removed latent channels tend to have positive scores: 54\% have $s_j>0$,
compared with 15\% of retained channels. Furthermore, the gate removes 89\% of the top 10\% most background-associated channels.
These findings suggest that semantic-segmentation guidance encourages the gate to
selectively suppress background-associated latent channels.
}

\section{Conclusion}
\label{sec:conclusion}

This paper presents a method that bridges modular and end-to-end autonomous driving by using perception modules as training-time teachers to compress the visual representations of end-to-end driving models.
Specifically, \emph{semantic-segmentation guidance} uses a gated autoencoder to remove driving-irrelevant background components inside image patch tokens, while \emph{object-detection guidance} contrastively pulls together scene-level register tokens across driving scenes that share similar traffic layouts, routes, and future trajectories.
Evaluated across six end-to-end driving models spanning ViT and ResNet backbones on the NAVSIM benchmarks, our approach consistently improves out-of-distribution driving performance without requiring any perception modules at inference.
We hope these findings inspire further research at the intersection of structured perception and end-to-end learning toward more robust and generalizable embodied AI.

\section*{LLM Usage Statement}

We used the LLM-based coding assistants Codex (OpenAI) and Claude Code
(Anthropic) in this work.
In the research, they assisted with coding, including the implementation,
training, and evaluation code.
In the writing, they assisted with grammar, \LaTeX{} formatting, drafting
text, and creating figure illustrations.
The authors checked all LLM-generated output, including code, results, text,
and figures, and take full responsibility for the content of this paper.

{
    \small
    \bibliographystyle{ieeenat_fullname}
    \bibliography{main}
}

\clearpage
\maketitlesupplementary
\section{Method Details}
\label{sec:supp-method}

This section provides the full formulations and training details of \Cref{sec:method} for reproducibility.

\paragraph{Autoencoder pre-fitting.}
In \Cref{sec:seg-teacher}, a linear autoencoder with encoder weights $\mathbf{W}_{\mathrm{enc}}\in\mathbb{R}^{m\times d}$, decoder weights $\mathbf{W}_{\mathrm{dec}}\in\mathbb{R}^{d\times m}$, and biases $\mathbf{b}_{\mathrm{enc}}\in\mathbb{R}^{m}$ and $\mathbf{b}_{\mathrm{dec}}\in\mathbb{R}^{d}$ decomposes each patch token $\mathbf{x}_i\in\mathbb{R}^d$ into $m>d$ latents:
\begin{equation}
\begin{aligned}
    \mathbf{z}_i &=
    \mathbf{W}_{\mathrm{enc}}(\mathbf{x}_i-\mathbf{b}_{\mathrm{dec}})
    + \mathbf{b}_{\mathrm{enc}},
    \\
    \hat{\mathbf{x}}_i &=
    \mathbf{W}_{\mathrm{dec}}\mathbf{z}_i+\mathbf{b}_{\mathrm{dec}}.
    \label{eq:ae}
\end{aligned}
\end{equation}
Each column of $\mathbf{W}_{\mathrm{dec}}$ is normalized to unit norm so the code magnitude $|z_{i,j}|$ measures the contribution of latent $j$ to token $i$ and the $\ell_1$ sparsity penalty cannot be bypassed by rescaling the decoder.
The autoencoder is trained with
\begin{equation}
    \mathcal{L}_{\mathrm{gate\_dict}} =
    \underbrace{\frac{\sum_i\|\hat{\mathbf{x}}_i-\mathbf{x}_i\|_2^2}
    {\sum_i\|\mathbf{x}_i-\bar{\mathbf{x}}\|_2^2+\epsilon}}
    _{\mathcal{L}_{\mathrm{rec}}}
    + \lambda_1
    \underbrace{\frac{1}{Nm}\sum_i\|\mathbf{z}_i\|_1}
    _{\mathcal{L}_{1}},
    \label{eq:ae-loss}
\end{equation}
where $\bar{\mathbf{x}}$ is the batch mean of the $N$ patch tokens, $\epsilon=10^{-8}$ is a numerical stability constant, and $\lambda_1$ weights the $\ell_1$ sparsity loss $\mathcal{L}_1$.
Before the joint training stage of \Cref{sec:training-objective}, we pre-fit the autoencoder on patch tokens from the frozen, pretrained image encoder prior to training.
Pre-fitting uses $8.2\times 10^{5}$ tokens ($512$ per frame from $1{,}600$ \texttt{navtrain} frames for ViT base models, and $256$ per frame from $3{,}200$ frames for ResNet-34) with Adam (learning rate $10^{-3}$, batch size $8{,}192$, $8{,}000$ steps, or $12{,}000$ for Drive-JEPA). We keep the checkpoint with the lowest reconstruction error on held-out frames.
During joint training, $\mathcal{L}_{\mathrm{gate\_dict}}$ keeps training the autoencoder but does not update the image encoder (we stop gradients on $\mathbf{x}_i$ in \Cref{eq:ae-loss}).

\paragraph{Gate schedule and hard gate.}
The gate regularization $\mathcal{L}_{\mathrm{gate\_reg}}$ of \Cref{sec:seg-teacher} has two terms: a compression loss $\mathcal{L}_{\mathrm{budget}}$ (its first term) and a binarization loss $\mathcal{L}_{\mathrm{bin}}$ (its second term).
$\mathcal{L}_{\mathrm{budget}}$ drives the mean gate activation toward a target keep ratio $\rho_t$ annealed linearly from $1$ to $\rho=0.6$ between $10\%$ and $40\%$ of training, while $\mathcal{L}_{\mathrm{bin}}$ pushes each gate toward $0$ or $1$.
Since $\mathcal{L}_{\mathrm{budget}}$ alone could be satisfied by setting every gate to $\rho$, we enable $\mathcal{L}_{\mathrm{bin}}$ once $\rho_t$ reaches $\rho$, ramping its weight linearly from zero over the following $30\%$ of training.
In the final quarter of training, we switch from the soft gate $g_j^{\mathrm{s}}=\sigma(a_j)/\sigma(a^{0})$ to the binary gate $g_j^{\mathrm{h}}=\mathbf{1}\{\sigma(a_j)>0.5\}$ with a straight-through estimator:
\begin{equation}
    g_j = g_j^{\mathrm{s}} +
    \operatorname{sg}\!\left(g_j^{\mathrm{h}}-g_j^{\mathrm{s}}\right),
    \label{eq:ste}
\end{equation}
where $\operatorname{sg}$ denotes stop-gradient.
In \Cref{eq:subtractive-gate}, $\mathbf{1}-\mathbf{g}$ is clamped to $[0,1]$, so a gate logit above $a^{0}$ leaves its latent unchanged. At inference, the binary gate $g_j^{\mathrm{h}}$ is used, so the module is a fixed affine map of each patch token.

\paragraph{Removal score.}
The segmentation pseudo-labels are grouped into seven classes: road and markings, vehicles and people, signs and lights, sky, vegetation, buildings, and other.
Each patch takes its majority class. Patches whose majority class covers less than half of their pixels are labeled \emph{other} and enter neither $\mathcal{L}_{\mathrm{seg\_rm}}$ nor $\mathcal{L}_{\mathrm{seg\_pre}}$. The head $h$ is a single linear layer with one output per class.
Let $y_i$ denote the semantic class label of patch $i$.
For each token $i$, we normalize the code magnitudes across the $m$ latents as $\tilde{z}_{i,j}=|z_{i,j}|/\big(\sum_{k=1}^{m}|z_{i,k}|+\epsilon\big)$ so each patch contributes its relative share independent of the token norm.
The class means in \Cref{eq:removal-score} are $\bar{z}^{\mathcal{R}}_j=\operatorname{mean}_{i:\,y_i\in\mathcal{R}}\tilde{z}_{i,j}$ and $\bar{z}^{\mathcal{U}}_j=\operatorname{mean}_{i:\,y_i\in\mathcal{U}}\tilde{z}_{i,j}$, tracked with an exponential moving average over training batches.
Centering by the median gives a positive score to the half of the latents that leans most toward the driving-irrelevant classes regardless of the class balance.
In \Cref{eq:rm-loss}, $s_j$ is detached so that $\mathcal{L}_{\mathrm{seg\_rm}}$ updates only the gate logits.

\paragraph{Loss weights.}
DrivoR and Drive-JEPA use the loss weights of \Cref{sec:training-objective}, with the two terms of $\mathcal{L}_{\mathrm{gate\_reg}}$ weighted equally.
On the ResNet-34 base models, the gate penalties are scaled to each base model's driving loss. $\mathcal{L}_{\mathrm{seg\_rm}}$ and $\mathcal{L}_{\mathrm{gate\_reg}}$ are weighted by $57$ and $114$ for LTF and MeanFuser, and by $200$ and $400$ for DiffusionDrive and PRIX.

\paragraph{BEV traffic map.}
The detection teacher encodes two inputs. The first is the future trajectory $\mathbf{T}\in\mathbb{R}^{8\times 3}$, which holds the ego poses $(x,y,\psi)$ at $0.5$\,s intervals. The second is the polar BEV traffic map $\mathbf{M}$ of \Cref{sec:contrastive}.
The map is first rasterized in the Cartesian frame at $0.25$\,m resolution, with the navigation route and the future trajectory drawn as $0.5$\,m-wide lines. Each cell of the polar frame, with uniform range and bearing bins, then stores the fraction it covers.
Both $\mathbf{T}$ and $\mathbf{M}$ are derived from the annotations, maps, and logs of \texttt{navtrain} and used only during training.
The important-object strip test is applied to the current 3D detection boxes and, for vehicles, pedestrians, and bicycles, to their boxes over the next $4$\,s. Objects whose bounding box comes within $1$\,m of the ego vehicle within $4$\,s are also labeled important. When the ego vehicle is nearly stationary, the strip is extended straight ahead to at least $15$\,m.

\paragraph{Detection teacher.}
The map encoder has four $3\times 3$ convolutions of stride $2$ with $16$, $32$, $64$, and $64$ channels, followed by a linear layer. Its output and the flattened $\mathbf{T}$ feed an MLP of widths $536$--$512$--$256$--$32$ that outputs $\mathbf{z}^{\mathrm{t}}$, normalized to $\hat{\mathbf{z}}^{\mathrm{t}}=\mathbf{z}^{\mathrm{t}}/\|\mathbf{z}^{\mathrm{t}}\|_2$.
From the unnormalized $\mathbf{z}^{\mathrm{t}}$, an MLP of widths $32$--$256$--$512$--$24$ reconstructs the poses $\hat{\mathbf{T}}$. A decoder outputs the map logits $\hat{\mathbf{M}}$ with a linear projection to $64\times 4\times 4$, three $4\times 4$ transposed convolutions of stride $2$ with $32$, $32$, and $16$ channels, and a $3\times 3$ convolution. All hidden layers use GELU activations.

\paragraph{Detection teacher pretraining.}
The detection teacher is trained with the binary cross-entropy loss $\mathcal{L}_{\mathrm{bce}}$ and the $\ell_1$ loss $\mathcal{L}_{\ell_1}$ of \Cref{sec:training-objective}, and two auxiliary terms:
\begin{equation}
    \mathcal{L}_{\mathrm{teacher}}
    =\mathcal{L}_{\mathrm{bce}}+\mathcal{L}_{\ell_1}
    +\mathcal{L}_{\mathrm{disp}}+3\,\mathcal{L}_{\mathrm{dist}},
    \label{eq:teacher-loss}
\end{equation}
where $\mathcal{L}_{\ell_1}=\frac{1}{8}\sum_{k=1}^{8}\|\hat{\mathbf{T}}_k-\mathbf{T}_k\|_1$, $\mathcal{L}_{\mathrm{disp}}$ applies the same norm to the displacements between consecutive poses, and
\begin{equation}
\begin{aligned}
    \mathcal{L}_{\mathrm{bce}}
    &=\frac{\sum_{c} w_c\,\mathrm{BCE}\big(\hat{\mathbf{M}}^{c},\mathbf{M}^{c}\big)}{\sum_{c} w_c}, \\
    \mathcal{L}_{\mathrm{dist}}
    &=\operatorname*{mean}_{a\neq b}\;
    \ell_{\mathrm{s}}\big(1-S^{\mathrm{t}}_{ab}-q_{ab}\big).
\end{aligned}
\label{eq:teacher-terms}
\end{equation}
Here, $\mathrm{BCE}$ is the binary cross-entropy of channel $c$ against the covered fractions, $\ell_{\mathrm{s}}$ is the smooth-$\ell_1$ penalty, and the mean runs over all pairs of distinct scenes in a batch.
The target $q_{ab}$ combines trajectory and map distances:
\begin{equation}
    d_{ab} = \frac{d^{\mathrm{traj}}_{ab}}{\bar{d}^{\mathrm{traj}}}
           + \frac{d^{\mathrm{map}}_{ab}}{\bar{d}^{\mathrm{map}}},
    \qquad
    q_{ab} = \frac{d_{ab}/\bar{d}}{1 + d_{ab}/\bar{d}}\in[0,1),
    \label{eq:pairs}
\end{equation}
where $d^{\mathrm{traj}}_{ab}$ is the mean $\ell_2$ distance between the future $(x,y)$ waypoints and $d^{\mathrm{map}}_{ab}=\sum_c w_c\,\delta^{c}_{ab}/\sum_c w_c$. Here $\delta^{c}_{ab}$ is the mean squared difference between channel $c$ of the two maps after $4\times 4$ average pooling, and bars denote means over the distinct pairs of the batch.
We set $w_c\propto 1/s_c$, with $s_c$ the coverage share of channel $c$, and $w_c=0$ for the four road channels (\Cref{sec:map_teacher} compares alternative weightings).
The detection teacher is trained on \texttt{navtrain} for $60$ epochs with AdamW (learning rate $10^{-3}$, weight decay $10^{-4}$, batch size $256$, gradient-norm clipping at $1$). The learning rate warms up linearly over the first $10\%$ of steps and then follows cosine decay. We keep the final checkpoint.
On held-out \texttt{navtrain} scenes, it reconstructs $\mathbf{T}$ with an average displacement error of $0.02$\,m, and its cosine distances $1-S^{\mathrm{t}}_{ab}$ correlate with $d_{ab}$ at Pearson $r=0.79$.

\paragraph{Pair sampling.}
We maintain a first-in-first-out queue of $512$ past detached mean register tokens $\mathbf{e}$ paired with their teacher latents $\hat{\mathbf{z}}^{\mathrm{t}}$.
For each anchor, we draw $64$ pairs uniformly from the current batch and the queue, excluding the anchor itself.
We set the margin hyperparameters of \mbox{\Cref{eq:contrastive}} to $\mu=0.4$ and $\gamma=1$.

\paragraph{Register tokens and ResNet-34 base models.}
DrivoR natively provides $16$ register tokens.
On Drive-JEPA, the Det.\ and Seg.\ + Det.\ variants add register tokens. They are concatenated to the patch tokens after the patch embedding and pass through all encoder blocks. The planner queries read them before each refinement step.
ResNet-34 base models have no register tokens, so $\mathcal{L}_{\mathrm{det}}$ acts on the mean-pooled spatial features that the policy head also reads. With Det.\ alone, its gradient changes the image encoder, which does not help on any of the four ResNet-34 base models. In Seg.\ + Det., we therefore stop this gradient at the encoder features, so it updates only the gate, autoencoder, and projection head.

\subsection{Detection Teacher and Road-Layout Ablation}
\label{sec:map_teacher}

\begin{figure*}[t]
  \centering
  \includegraphics[width=\textwidth]{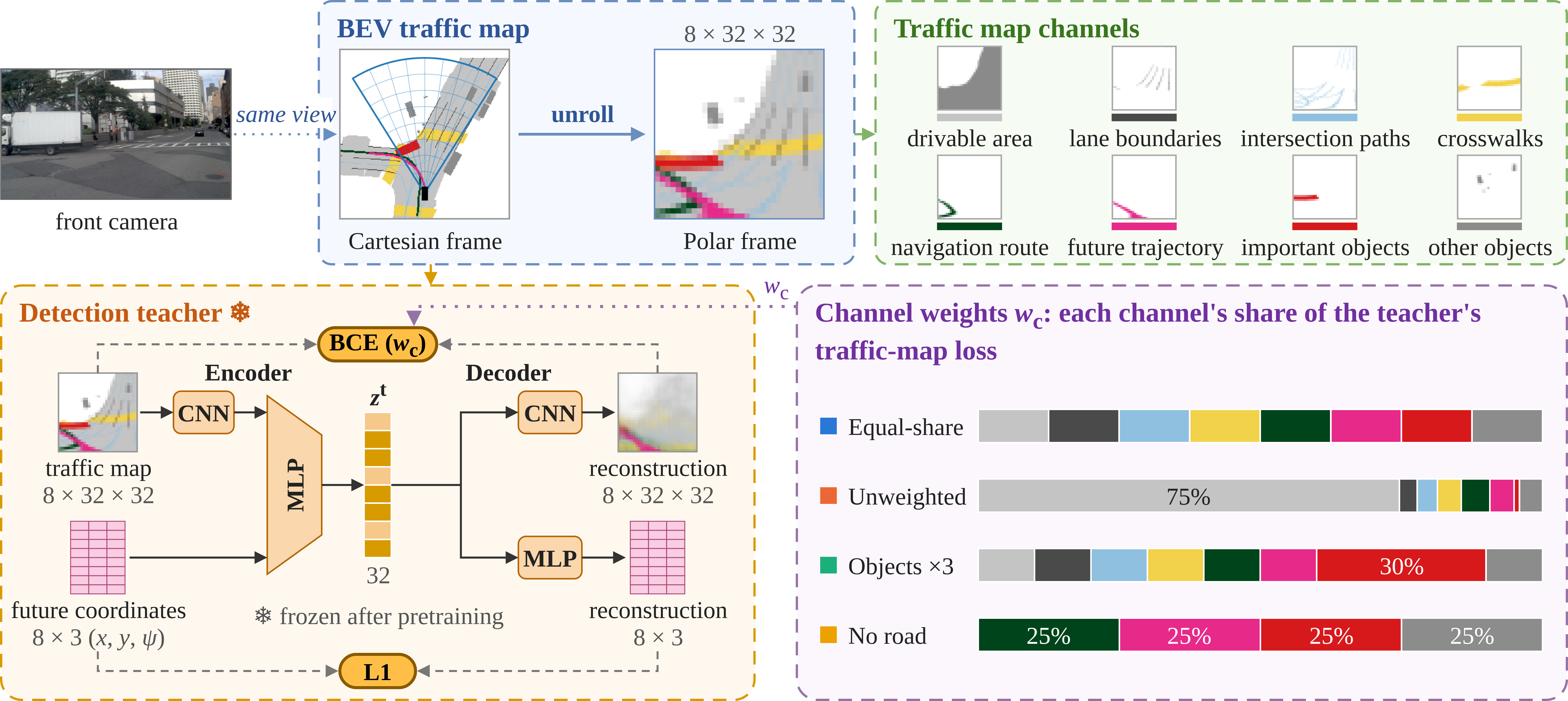}
  \caption{Object-detection guidance: the detection teacher. Top: the BEV traffic map is unrolled from the Cartesian frame into a $32 \times 32$ polar frame only within the field of view of the camera frustum. Middle: the traffic map channels, the teacher inputs. Bottom: the detection teacher encodes the polar BEV traffic map and the future trajectory into the latent $\mathbf{z}^{\mathrm{t}}$ and is frozen after pretraining. Used during training only.}
  \label{fig:map_teacher_mini}
\end{figure*}

This ablation varies the channel weights $w_c$ of the padded eight-channel polar BEV traffic map (\cref{fig:map_teacher_mini}), which set the contribution of each channel to $\mathcal{L}_{\mathrm{bce}}$ in \Cref{eq:teacher-terms} and to the target distance $d_{ab}$ in \Cref{eq:pairs}. The teacher is otherwise unchanged.

\paragraph{Conditions.}
The channel weights $w_c$ define which spatial cues count toward driving-condition similarity.
Raw coverage is dominated by the drivable area (75\%). Every other channel covers $3$--$5\%$, except important objects ($1\%$).
We compare four weightings. \textbf{Unweighted} sets $w_c = 1$. \textbf{Equal-share} sets $w_c \propto 1/s_c$, where $s_c$ is the coverage share of the channel. \textbf{Objects $\times$3} is equal-share with important objects tripled. \textbf{No road} is our main detection teacher and gives zero weight to the four road channels.
The control is \textbf{No teacher}, trained with the driving loss only.

\Cref{tab:teacher-weighting} reports last-epoch navhard EPDMS for front-camera DrivoR. Only No road beats the control ($47.7$ vs.\ $46.3$). Equal-share matches it, and Unweighted and Objects $\times$3 fall below it.

\begin{table}[t]
\centering
\caption{Detection-teacher channel weighting.
Navhard EPDMS at the last training epoch of front-camera DrivoR, for each channel weighting of the detection teacher
(\Cref{sec:map_teacher}, \Cref{fig:map_teacher_mini}). The control is trained on the driving loss only. The best
value is in bold.}
\label{tab:teacher-weighting}
\begin{tblr}{
  colspec = {l c},
  row{1}  = {font=\bfseries},
}
\toprule
Teacher              & EPDMS          \\
\midrule
Equal-share          & 46.3          \\
Unweighted           & 44.9          \\
Objects $\times$3    & 45.8          \\
No road              & \textbf{47.7} \\
\midrule
No teacher (control) & 46.3          \\
\bottomrule
\end{tblr}
\end{table}

\section{Experimental Details}

\subsection{Benchmarks and Evaluation Protocol}
\label{app:benchmark-protocol}

\paragraph{NAVSIM-v1.}
To verify that robustness improvements do not compromise
nominal driving, we evaluate the Predictive Driver Model Score
(PDMS) on the \texttt{navtest} split of
NAVSIM-v1~\cite{dauner2024navsim}.

\paragraph{NAVSIM-v2.}
Our primary benchmark is NAVSIM-v2~\cite{cao2025pseudosimulation},
built on nuPlan and evaluated through pseudo-simulation.
The \texttt{navhard-two-stage} split first evaluates a driving
policy on real observations and subsequently on views rendered
using 3D Gaussian Splatting from perturbed ego poses.
This two-stage evaluation tests both nominal driving
and recovery from states outside the expert distribution.

We report the Extended Predictive Driver Model Score (EPDMS) using the official NAVSIM-v2 evaluator. For DrivoR, proposals are ranked using the NAVSIM-v2
scoring weights:
$w_{\mathrm{NC}}=10$,
$w_{\mathrm{DAC}}=13$,
$w_{\mathrm{DDC}}=6$,
$w_{\mathrm{TTC}}=14$,
$w_{\mathrm{EP}}=15$, and
$w_{\mathrm{comfort}}=2$.

To validate our scoring pipeline, we evaluate the released
four-camera DrivoR checkpoint and obtain an EPDMS of 48.31,
consistent with the 48.3 reported in~\cite{kirby2026driving}.

\paragraph{Metrics.}
EPDMS~\cite{cao2025pseudosimulation} combines four multiplier metrics with five weighted metrics. The multiplier metrics are no at-fault collisions (NC), drivable area compliance (DAC), driving direction compliance (DDC), and traffic light compliance (TLC). The weighted metrics are ego progress (EP), time to collision (TTC), lane keeping (LK), history comfort (HC), and extended comfort (EC):
\begin{equation}
\begin{aligned}
    \mathrm{EPDMS}={}&\mathrm{NC}\cdot\mathrm{DAC}\cdot\mathrm{DDC}\cdot\mathrm{TLC}\\
    &\cdot\frac{5\,\mathrm{EP}+5\,\mathrm{TTC}+2\,\mathrm{LK}+2\,\mathrm{HC}+2\,\mathrm{EC}}{16},
\end{aligned}
\label{eq:epdms}
\end{equation}
where a metric is set to $1$ in scenes in which the human driver also violates it. On \texttt{navhard-two-stage}, the Stage~2 scores of a scene are averaged and the result is multiplied by the Stage~1 score. The averaging weights decay with the distance between the Stage~2 start pose and the Stage~1 end pose of the evaluated policy. PDMS on NAVSIM-v1~\cite{dauner2024navsim} is $\mathrm{NC}\cdot\mathrm{DAC}\cdot(5\,\mathrm{EP}+5\,\mathrm{TTC}+2\,\mathrm{Comf.})/12$, with Comf.\ the comfort metric.

\paragraph{Reporting protocol.}
No checkpoints are selected using the test split.
Unless otherwise specified, each reported NAVSIM-v2 result
corresponds to the final training epoch and is expressed
as the mean and standard deviation across five random seeds.

Each base model and its corresponding baseline use identical
training recipes to ensure controlled comparisons.

We observe that NAVSIM-v2 navhard EPDMS fluctuates by
approximately $\pm1$ point across runs and by as much as
$\pm2$ points across epochs within a single run.
Consequently, single-run differences below one EPDMS point
are treated as evaluation noise rather than conclusive
performance improvements.

\subsection{Implementation and Training Details}
\label{app:implementation-details}

\paragraph{DrivoR.}
Our primary base model is DrivoR~\cite{kirby2026driving},
equipped with a DINOv2 ViT-S/14 visual backbone and
rank-32 LoRA adaptation.
DrivoR processes the current front-camera frame, resized to $1148\times 672$ ($48\times 82$ patch tokens).

Both the baseline and our modified model are trained on
\texttt{navtrain} for 10 epochs using the official
DrivoR NAVSIM-v2 training recipe.
Optimization uses AdamW with an initial learning rate of
$2\times10^{-4}$, batch size 64, and a learning-rate schedule
consisting of a 10\% linear warm-up followed by cosine decay.

\paragraph{Additional base models.}
To examine generality across different visual representations,
we integrate our module into Drive-JEPA~\cite{wang2026drivejepa}
and four ResNet-34-based base models:
Latent TransFuser (LTF)~\cite{chitta2023transfuser},
DiffusionDrive~\cite{liao2025diffusiondrive},
PRIX~\cite{wozniak2025prix}, and
MeanFuser~\cite{wang2026meanfuser}.

For Drive-JEPA, the module is inserted after visual
encoder block 22.
Drive-JEPA encodes the front-camera frames at $t-1$ and $t$ ($512\times 256$) with a ViT-L of $24$ blocks into $16\times 32$ patch tokens of dimension $d=1024$, giving $m=4096$ latents.
Training follows the authors' official recipe for 20 epochs (Adam, learning rate $10^{-4}$, scaled by $0.1$ for the encoder, batch size $64$).

For the four ResNet-based base models, the visual input consists
of stitched front-left, front, and front-right camera views
from the current frame.
The module operates on the $8\times32$ spatial tokens
extracted after the final ResNet stage.
Each base model is trained for 100 epochs using its
corresponding official training recipe.
All four are run camera-only. LTF uses Adam (learning rate $10^{-4}$, batch size $64$) with a cosine schedule. DiffusionDrive and PRIX use AdamW (learning rate $6\times 10^{-4}$, warm-up and cosine decay, effective batch size $512$). MeanFuser uses AdamW (learning rate $2\times 10^{-4}$, warm-up and cosine decay, batch size $32$).

\paragraph{Latents and gating.}
For DrivoR, the module is inserted after DINOv2 block 10,
where the feature dimension is $d=384$.
Blocks are indexed from $0$, so the module acts on the output of the penultimate block of both ViT encoders.
The autoencoder has $m=1536$ latents.
For the ResNet-based base models, the number of latents is
increased to $m=2048$.

We employ the subtractive gate defined in
\Cref{eq:subtractive-gate}, initializing all gate logits
to $a^0=2$.

Gate logits are optimized with a learning rate of
$6\times10^{-3}$ and no weight decay.

\paragraph{Segmentation and removal supervision.}
Segmentation pseudo-labels are generated offline using
Mask2Former with a Swin-L backbone pretrained on
Mapillary Vistas~\cite{cheng2022masked}.
Labels are predicted per camera and pooled to the token grid of each base model ($48\times 82$ for DrivoR, $16\times 32$ for Drive-JEPA, and $8\times 32$ on the stitched $1024\times 256$ view for ResNet-34).

The removal loss $\mathcal{L}_{\mathrm{seg\_rm}}$ is activated
after 20 optimization steps, allowing the moving averages
defined in \Cref{eq:removal-score} to warm up.
These moving averages use a decay factor of 0.99.

\begin{table*}[t]
  \centering
  \footnotesize
  \caption{Compute overhead of our module on DrivoR.
  Inference at batch size 1 in fp32 and training at batch size 64 in mixed precision, both on one A100.
  Parentheses give the overhead relative to the baseline.
  Training-only modules (segmentation head, detection teacher) are excluded from the inference columns.}
  \label{tab:compute}
  \setlength{\tabcolsep}{0pt}
  \begin{tabular*}{\textwidth}{@{\extracolsep{\fill}}l cccc cccc@{}}
    \toprule
    & \multicolumn{4}{c}{Inference (batch 1)} & \multicolumn{4}{c}{Training (batch 64)} \\
    \cmidrule(lr){2-5}\cmidrule(lr){6-9}
    Configuration & Params (M) & GFLOPs & Latency (ms) & Memory (GB) & Trainable (M) & Step (s) & Memory (GB) & GPU-h (10 ep.) \\
    \midrule
    DrivoR        & 40.8          & 465          & 34.1          & 0.25          & 17.7          & 4.60          & 34.3          & 17.0          \\
    + Seg.        & 42.0 (+2.9\%) & 479 (+3.0\%) & 34.9 (+2.5\%) & 0.27 (+11\%)  & 18.9 (+6.7\%) & 4.84 (+5.1\%) & 44.3 (+29\%)  & 17.9 (+5.1\%) \\
    + Det.        & 40.8 (+0.0\%) & 465 (+0.0\%) & 34.1 (+0.0\%) & 0.25 (+2.4\%) & 17.7 (+0.1\%) & 4.93 (+7.2\%) & 34.3 (+0.0\%) & 18.2 (+7.2\%) \\
    + Seg. + Det. & 42.0 (+2.9\%) & 479 (+3.0\%) & 35.0 (+2.6\%) & 0.28 (+13\%)  & 18.9 (+6.8\%) & 4.85 (+5.4\%) & 44.3 (+29\%)  & 17.9 (+5.4\%) \\
    \bottomrule
  \end{tabular*}
\end{table*}

\paragraph{Object-detection guidance.}
The contrastive projection head is optimized with a learning rate of $5\times10^{-3}$.

\paragraph{Compute overhead.}
\Cref{tab:compute} measures the cost of our module on front-camera DrivoR for the four DrivoR configurations of \Cref{fig:planners}: the baseline, segmentation guidance, detection guidance, and both.
All configurations are measured on one A100-80GB with the same input resolution, the trained checkpoint of each configuration, and the same real \texttt{navtrain} samples.
Inference is measured in evaluation mode at batch size 1 in fp32 (TF32 off) over 200 timed forward passes after 20 warm-up passes, counting only the modules that execute at inference. The segmentation head and the detection teacher are used only in the training loss and are excluded.
Training is measured as the mean of 30 full optimizer steps (forward, all losses, backward, update) at the training batch size 64 in mixed precision, as in all our runs. GPU-hours extrapolate this step time to the 13{,}290 steps of a 10-epoch run and exclude data loading.
The gated autoencoder adds 1.18M parameters (two $384\times1536$ matrices with biases and gate logits) and about 3\% to the inference FLOPs and latency.
The detection teacher adds no inference cost because only the register tokens it supervises remain in the model.
During training, segmentation guidance raises peak memory by 29\% because the $m$-dimensional latents of every patch are kept for the backward pass, and either teacher adds 5--7\% to the step time.

\subsection{Ablation and Analysis Details}
\label{app:analysis-details}

\paragraph{Token-masking baselines.}
The baselines of \Cref{tab:removal} use the Mask2Former labels of the segmentation teacher: $f_i$ is the fraction of pixels of patch $i$ labeled with a class in $\mathcal{U}$.
\emph{Input} scales the output of the patch embedding, and \emph{Block~10} scales the patch tokens at the output of block $10$, where our gated autoencoder acts. Register tokens are left unchanged.
\emph{Random} draws a new set of $\operatorname{round}(\sum_i f_i)$ patch tokens for every frame.
The masks are applied during training and at test time, where Mask2Former is run on the test images. The model and training recipe are otherwise those of the DrivoR baseline.

\paragraph{Driving-condition retrieval.}
The $3{,}000$ scenes of \Cref{tab:condition-retrieval} are sampled uniformly at random from \texttt{navtest}, and features are extracted with the last-epoch checkpoints of one seed.
Teacher latents are computed from the \texttt{navtest} annotations and logs.
All similarities are cosine similarities, and the rank correlation is computed over all pairs of scenes from different drives.
Neighbors are retrieved with at most one scene per drive, the trajectory distance is the mean $\ell_2$ distance between the eight future waypoints, and the Random row averages over pairs of scenes from different drives.

\paragraph{Latent channel selection.}
For \Cref{fig:removal}, the removal scores $s_j$ are recomputed on $200$ \texttt{navtrain} frames with a last-epoch checkpoint, and a channel counts as removed if its binary gate is closed.

\section{NAVSIM-v2 Sub-Metrics}
\label{sec:supp-navsim-v2}

\Cref{tab:supp-planners} lists the per-metric \texttt{navhard-two-stage} scores behind \Cref{fig:planners} for every base model and variant. The standard deviation of DrivoR with Seg.\ + Det.\ ($2.3$) is driven by one seed at $43.9$ EPDMS. The other four score between $46.6$ and $49.9$.

\begin{table*}[!t]
    \fycolor
    \centering
    \caption{NAVSIM-v2 sub-metrics on \texttt{navhard-two-stage} for the runs of \Cref{fig:planners} (last epoch, five seeds unless marked). EPDMS is the mean with the standard deviation in parentheses. Cyan and orange mark the highest and second-highest EPDMS within each base model (ranked before rounding). Each sub-metric cell gives Stage~1 (real observations)\,/\,Stage~2 (synthetic observations), averaged over the seeds. EPDMS multiplies the four multiplier metrics (NC, DAC, DDC, TLC) by the weighted average of the weighted metrics EP (5), TTC (5), LK (2), HC (2), and EC (2). Top: ViT-based base models. Bottom: ResNet-34 base models. Seg.: semantic-segmentation guidance with the gated autoencoder (\Cref{sec:seg-teacher}). Det.: object-detection guidance (\Cref{sec:contrastive}), alone or jointly with it. On Drive-JEPA, Det.\ acts on 16 added register tokens.}
    \label{tab:supp-planners}
    \footnotesize
    \setlength{\tabcolsep}{0pt}
    \definecolor{tabBest}{RGB}{0,145,170}
    \definecolor{tabSecond}{RGB}{225,120,30}
    \newcommand{\sd}[1]{\,{\scriptsize(#1)}}
    \newcommand{\first}[2]{\textcolor{tabBest}{\textbf{#1}\sd{#2}}}
    \newcommand{\second}[2]{\textcolor{tabSecond}{\textbf{#1}\sd{#2}}}
    \begin{tabular*}{\textwidth}{@{\extracolsep{\fill}}ll cccc ccccc c@{}}
        \toprule
        & & \multicolumn{4}{c}{Multipliers ($\times$), Stage 1 / Stage 2} & \multicolumn{5}{c}{Weighted, Stage 1 / Stage 2} & \\
        \cmidrule(lr){3-6}\cmidrule(lr){7-11}
        Base model & Variant & NC$\uparrow$ & DAC$\uparrow$ & DDC$\uparrow$ & TLC$\uparrow$ & EP$\uparrow$ & TTC$\uparrow$ & LK$\uparrow$ & HC$\uparrow$ & EC$\uparrow$ & EPDMS$\uparrow$ \\
        \midrule
        DrivoR~\cite{kirby2026driving} & Baseline & 99.0\,/\,88.9 & 94.7\,/\,87.7 & 98.8\,/\,92.1 & 100.0\,/\,98.6 & 73.3\,/\,67.6 & 99.0\,/\,87.0 & 92.6\,/\,52.6 & 97.6\,/\,98.5 & 69.8\,/\,76.7 & 46.3\sd{1.0} \\
         & Seg. & 98.7\,/\,88.9 & 94.9\,/\,87.5 & 98.9\,/\,92.9 & 99.9\,/\,99.0 & 73.5\,/\,68.3 & 98.8\,/\,86.9 & 93.5\,/\,52.7 & 97.6\,/\,98.4 & 67.2\,/\,76.7 & 46.7\sd{1.4} \\
         & Det. & 99.0\,/\,89.5 & 94.9\,/\,89.7 & 99.1\,/\,92.8 & 99.9\,/\,98.7 & 72.7\,/\,65.3 & 98.7\,/\,87.7 & 93.8\,/\,54.1 & 97.6\,/\,98.9 & 67.9\,/\,77.3 & \first{47.7}{0.2}$^3$ \\
         & Seg.\ + Det. & 98.9\,/\,89.0 & 95.5\,/\,89.0 & 99.0\,/\,92.9 & 99.7\,/\,98.7 & 72.9\,/\,67.8 & 98.9\,/\,86.7 & 93.7\,/\,52.4 & 97.6\,/\,98.6 & 68.2\,/\,75.5 & \second{47.3}{2.3} \\
        \addlinespace[2pt]
        Drive-JEPA~\cite{wang2026drivejepa} & Baseline & 93.1\,/\,82.4 & 93.0\,/\,84.0 & 96.3\,/\,85.4 & 99.6\,/\,98.7 & 74.4\,/\,51.9 & 92.3\,/\,77.1 & 98.0\,/\,54.6 & 97.1\,/\,93.9 & 37.8\,/\,24.5 & 32.2\sd{0.4}$^3$ \\
         & Seg. & 93.4\,/\,82.7 & 93.2\,/\,84.2 & 96.9\,/\,86.0 & 99.5\,/\,98.7 & 73.6\,/\,51.9 & 93.0\,/\,77.1 & 97.9\,/\,55.6 & 97.3\,/\,94.6 & 38.1\,/\,27.7 & 32.5\sd{1.7}$^3$ \\
         & Det. & 93.9\,/\,83.1 & 94.5\,/\,84.3 & 96.7\,/\,85.7 & 99.6\,/\,98.7 & 75.3\,/\,52.5 & 92.8\,/\,77.5 & 98.2\,/\,54.6 & 97.0\,/\,95.2 & 38.1\,/\,27.4 & \first{33.5}{1.7}$^3$ \\
         & Seg.\ + Det. & 92.9\,/\,82.8 & 94.4\,/\,84.2 & 96.6\,/\,86.1 & 99.6\,/\,98.8 & 73.9\,/\,52.3 & 92.4\,/\,77.0 & 98.6\,/\,55.4 & 97.2\,/\,94.2 & 36.3\,/\,23.7 & \second{32.7}{1.0}$^3$ \\
        \midrule
        LTF~\cite{chitta2023transfuser} & Baseline & 96.0\,/\,79.7 & 80.0\,/\,68.5 & 98.6\,/\,83.7 & 99.3\,/\,98.5 & 83.9\,/\,85.2 & 94.8\,/\,76.4 & 93.2\,/\,45.3 & 97.6\,/\,96.3 & 79.5\,/\,75.4 & 25.6\sd{0.9} \\
         & Seg. & 95.8\,/\,79.6 & 80.5\,/\,70.5 & 98.7\,/\,83.6 & 99.4\,/\,98.1 & 83.9\,/\,85.2 & 94.8\,/\,76.4 & 93.0\,/\,45.9 & 97.7\,/\,96.1 & 78.8\,/\,74.5 & \second{26.0}{1.1} \\
         & Det. & 96.3\,/\,79.1 & 79.7\,/\,67.6 & 98.5\,/\,82.7 & 99.4\,/\,98.5 & 83.8\,/\,85.6 & 94.4\,/\,75.8 & 93.4\,/\,45.1 & 97.6\,/\,95.8 & 79.3\,/\,74.0 & 24.6\sd{0.5} \\
         & Seg.\ + Det. & 96.5\,/\,79.5 & 80.0\,/\,69.3 & 98.9\,/\,83.8 & 99.6\,/\,98.5 & 83.8\,/\,85.7 & 95.2\,/\,76.0 & 94.6\,/\,45.8 & 97.7\,/\,96.2 & 79.0\,/\,75.0 & \first{26.0}{0.9} \\
        \addlinespace[2pt]
        MeanFuser~\cite{wang2026meanfuser} & Baseline & 96.5\,/\,79.9 & 88.3\,/\,74.7 & 99.1\,/\,86.5 & 99.4\,/\,98.3 & 84.4\,/\,86.8 & 95.5\,/\,76.6 & 96.4\,/\,48.3 & 97.6\,/\,96.1 & 79.7\,/\,69.9 & 32.5\sd{1.0} \\
         & Seg. & 96.9\,/\,80.2 & 88.1\,/\,75.0 & 99.3\,/\,86.8 & 99.3\,/\,98.3 & 84.3\,/\,86.8 & 95.9\,/\,76.6 & 96.5\,/\,49.2 & 97.6\,/\,96.4 & 79.6\,/\,70.9 & \first{33.0}{1.0} \\
         & Det. & 96.9\,/\,79.8 & 87.6\,/\,74.0 & 98.9\,/\,85.3 & 99.3\,/\,98.1 & 84.5\,/\,88.1 & 95.2\,/\,75.6 & 95.6\,/\,48.1 & 97.6\,/\,96.6 & 77.1\,/\,66.5 & 31.8\sd{1.0} \\
         & Seg.\ + Det. & 96.8\,/\,80.0 & 88.7\,/\,74.8 & 99.4\,/\,86.1 & 99.5\,/\,98.3 & 84.4\,/\,86.9 & 96.1\,/\,76.0 & 96.4\,/\,49.5 & 97.6\,/\,96.4 & 79.4\,/\,71.5 & \second{32.8}{0.5} \\
        \addlinespace[2pt]
        DiffusionDrive~\cite{liao2025diffusiondrive} & Baseline & 96.7\,/\,78.8 & 83.1\,/\,71.2 & 98.4\,/\,83.5 & 99.4\,/\,98.4 & 83.9\,/\,86.0 & 95.0\,/\,75.8 & 95.1\,/\,45.8 & 97.6\,/\,96.0 & 78.8\,/\,73.5 & 28.1\sd{1.1} \\
         & Seg. & 96.4\,/\,79.4 & 83.0\,/\,70.9 & 98.9\,/\,83.4 & 99.5\,/\,98.3 & 83.9\,/\,86.3 & 95.0\,/\,76.4 & 94.8\,/\,47.1 & 97.6\,/\,96.4 & 78.8\,/\,73.8 & \first{29.2}{1.7} \\
         & Det. & 96.4\,/\,79.4 & 81.4\,/\,70.5 & 98.7\,/\,83.0 & 99.4\,/\,98.4 & 84.0\,/\,85.9 & 94.8\,/\,76.7 & 94.7\,/\,45.5 & 97.6\,/\,96.3 & 78.5\,/\,73.9 & 27.2\sd{1.3} \\
         & Seg.\ + Det. & 96.6\,/\,79.6 & 83.0\,/\,70.2 & 98.6\,/\,83.2 & 99.4\,/\,98.3 & 84.1\,/\,85.9 & 94.8\,/\,76.4 & 95.1\,/\,46.4 & 97.6\,/\,96.4 & 79.0\,/\,73.9 & \second{28.2}{1.0} \\
        \addlinespace[2pt]
        PRIX~\cite{wozniak2025prix} & Baseline & 96.4\,/\,80.7 & 84.2\,/\,73.4 & 98.8\,/\,86.6 & 99.3\,/\,98.3 & 83.8\,/\,86.4 & 95.2\,/\,77.7 & 95.5\,/\,49.0 & 97.6\,/\,96.0 & 80.2\,/\,72.8 & 30.6\sd{2.1} \\
         & Seg. & 96.4\,/\,80.6 & 84.1\,/\,73.4 & 98.8\,/\,86.8 & 99.4\,/\,98.2 & 83.8\,/\,86.3 & 95.1\,/\,77.4 & 95.1\,/\,49.5 & 97.6\,/\,96.2 & 80.0\,/\,71.1 & \first{31.0}{1.2} \\
         & Det. & 96.3\,/\,80.1 & 84.5\,/\,73.3 & 98.6\,/\,86.3 & 99.3\,/\,98.1 & 83.9\,/\,86.5 & 95.1\,/\,76.3 & 95.7\,/\,49.9 & 97.6\,/\,96.3 & 78.7\,/\,72.1 & 30.5\sd{1.4} \\
         & Seg.\ + Det. & 96.2\,/\,80.0 & 84.5\,/\,73.6 & 99.0\,/\,86.5 & 99.3\,/\,98.1 & 83.8\,/\,86.7 & 94.5\,/\,76.9 & 95.8\,/\,49.5 & 97.6\,/\,96.3 & 79.6\,/\,71.7 & \second{30.7}{1.8} \\
        \bottomrule
    \end{tabular*}

    \smallskip
    \parbox{\textwidth}{\footnotesize $^{3}$Three seeds.}
\end{table*}

\section{NAVSIM-v1 Results}
\label{sec:supp-navsim-v1}

\Cref{tab:supp-navsim-v1} reports NAVSIM-v1 PDMS on \texttt{navtest} for the base models and variants of \Cref{fig:planners}, with the same checkpoints (last epoch) and seeds.
Unlike \texttt{navhard-\allowbreak two-\allowbreak stage}, \texttt{navtest} consists mostly of nominal scenes.

\begin{table*}[t]
  \centering
  \footnotesize
  \caption{NAVSIM-v1 results on \texttt{navtest} for the runs of \Cref{fig:planners} (last epoch, mean with standard deviation in parentheses over five seeds, three for Drive-JEPA). PDMS multiplies the two multiplier metrics (NC, DAC) by the weighted average of the weighted metrics TTC (5), Comf.\ (2), and EP (5). Cyan and orange mark the highest and second-highest PDMS within each base model (ranked before rounding). DrivoR is scored with its own NAVSIM-v1 evaluator. The other base models are scored with the NAVSIM-v2 code and the same PDMS formula.}
  \label{tab:supp-navsim-v1}
  \setlength{\tabcolsep}{0pt}
  \definecolor{tabBest}{RGB}{0,145,170}
  \definecolor{tabSecond}{RGB}{225,120,30}
  \newcommand{\sd}[1]{\,{\scriptsize(#1)}}
  \newcommand{\first}[2]{\textcolor{tabBest}{\textbf{#1}\sd{#2}}}
  \newcommand{\second}[2]{\textcolor{tabSecond}{\textbf{#1}\sd{#2}}}
  \begin{tabular*}{\textwidth}{@{\extracolsep{\fill}}llcccccc@{}}
    \toprule
    Base model & Variant & NC$\uparrow$ & DAC$\uparrow$ & TTC$\uparrow$ & Comf.$\uparrow$ & EP$\uparrow$ & PDMS$\uparrow$ \\
    \midrule
    DrivoR~\cite{kirby2026driving} & Baseline & 99.1 & 98.2 & 96.9 & 100.0 & 84.8 & 91.2\sd{0.2} \\
     & Seg. & 99.1 & 98.3 & 96.8 & 100.0 & 85.2 & \second{91.4}{0.2} \\
     & Det. & 99.1 & 98.3 & 96.8 & 100.0 & 85.0 & 91.3\sd{0.1} \\
     & Seg.\ + Det. & 99.1 & 98.3 & 96.9 & 100.0 & 85.3 & \first{91.5}{0.2} \\
    \addlinespace[2pt]
    Drive-JEPA~\cite{wang2026drivejepa} & Baseline & 97.5 & 98.1 & 96.8 & 97.7 & 88.9 & 92.1\sd{0.1} \\
     & Seg. & 97.7 & 97.9 & 96.9 & 97.9 & 88.6 & 91.9\sd{0.6} \\
     & Det. & 97.3 & 98.4 & 96.7 & 98.0 & 88.9 & \first{92.2}{0.7} \\
     & Seg.\ + Det. & 97.8 & 98.1 & 97.2 & 98.0 & 88.5 & \second{92.1}{0.5} \\
    \midrule
    LTF~\cite{chitta2023transfuser} & Baseline & 97.8 & 93.4 & 96.9 & 98.3 & 87.5 & 86.1\sd{0.3} \\
     & Seg. & 97.8 & 93.7 & 96.9 & 98.3 & 87.4 & \first{86.3}{0.4} \\
     & Det. & 97.8 & 93.3 & 96.9 & 98.3 & 87.5 & 86.0\sd{0.3} \\
     & Seg.\ + Det. & 97.8 & 93.4 & 97.0 & 98.3 & 87.4 & \second{86.1}{0.4} \\
    \addlinespace[2pt]
    MeanFuser~\cite{wang2026meanfuser} & Baseline & 97.9 & 96.8 & 97.1 & 98.3 & 87.7 & \first{89.3}{0.3} \\
     & Seg. & 97.9 & 96.7 & 97.0 & 98.3 & 87.8 & 89.2\sd{0.1} \\
     & Det. & 98.0 & 96.6 & 97.1 & 98.3 & 87.8 & 89.3\sd{0.2} \\
     & Seg.\ + Det. & 97.9 & 96.8 & 97.1 & 98.3 & 87.7 & \second{89.3}{0.2} \\
    \addlinespace[2pt]
    DiffusionDrive~\cite{liao2025diffusiondrive} & Baseline & 98.0 & 95.0 & 97.1 & 98.3 & 87.4 & 87.7\sd{0.2} \\
     & Seg. & 98.0 & 95.2 & 97.1 & 98.3 & 87.5 & \first{87.9}{0.3} \\
     & Det. & 98.1 & 94.7 & 97.2 & 98.3 & 87.5 & 87.5\sd{0.2} \\
     & Seg.\ + Det. & 98.0 & 95.1 & 97.1 & 98.3 & 87.5 & \second{87.8}{0.3} \\
    \addlinespace[2pt]
    PRIX~\cite{wozniak2025prix} & Baseline & 97.9 & 95.7 & 97.1 & 98.3 & 87.3 & \second{88.2}{0.2} \\
     & Seg. & 98.0 & 95.5 & 97.1 & 98.4 & 87.4 & 88.1\sd{0.2} \\
     & Det. & 97.9 & 95.8 & 97.1 & 98.4 & 87.4 & \first{88.4}{0.3} \\
     & Seg.\ + Det. & 97.9 & 95.6 & 97.1 & 98.4 & 87.5 & 88.2\sd{0.1} \\
    \bottomrule
  \end{tabular*}

  \vspace{12pt}
  \begin{minipage}[t]{\columnwidth}
    \centering
    \footnotesize
    \setlength{\tabcolsep}{0pt}
    \caption{Partial alignment teachers (navhard two-stage EPDMS at the last epoch, mean with standard deviation in parentheses over five seeds). Traj./road pairs: positive pairs defined by future trajectory alone or by road geometry alone, with the segmentation-guided gate.}
    \label{tab:supp-pairs}

    \begin{tabular*}{\columnwidth}{@{\extracolsep{\fill}}lccc@{}}
        \toprule
        Base model & Seg.\ + Det. & Seg.\ + traj.\ pairs & Seg.\ + road pairs \\
        \midrule
        DrivoR~\cite{kirby2026driving} & \textbf{47.3}\sd{2.3} & -- & 47.2\sd{0.4} \\
        LTF~\cite{chitta2023transfuser} & 26.0\sd{0.9} & \textbf{26.6}\sd{1.7} & 26.2\sd{1.1} \\
        MeanFuser~\cite{wang2026meanfuser} & 32.8\sd{0.5} & 33.2\sd{0.4} & \textbf{33.4}\sd{1.2} \\
        DiffusionDrive~\cite{liao2025diffusiondrive} & 28.2\sd{1.0} & 27.7\sd{1.3} & \textbf{28.4}\sd{1.1} \\
        PRIX~\cite{wozniak2025prix} & \textbf{30.7}\sd{1.8} & 30.6\sd{0.5} & \textbf{30.7}\sd{1.1} \\
        \bottomrule
    \end{tabular*}
\end{minipage}\hfill
  \begin{minipage}[t]{\columnwidth}
  \centering
  \footnotesize
  \caption{Navhard scenes in which the baseline or our module scores below 50 EPDMS (5{,}912 scenes in total), and how the two
  compare on them. Each scene's EPDMS is averaged over the five seeds of each model. Our module is as in \cref{fig:planners}
  (Seg.\ + Det.\ on DrivoR, Seg.\ on the ResNet base models). Ours higher or lower means by more than 5 EPDMS. The last three columns add up to the first.}
  \label{tab:supp-bev-pool}
  \setlength{\tabcolsep}{0pt}
  \begin{tabular*}{\columnwidth}{@{\extracolsep{\fill}}lcccc@{}}
    \toprule
    Base model & Below 50 & Ours higher & Ours lower & Within 5 \\
    \midrule
    DrivoR         & 2{,}081 & 306 & 228 & 1{,}547 \\
    LTF            & 2{,}974 & 518 & 393 & 2{,}063 \\
    MeanFuser      & 2{,}627 & 362 & 289 & 1{,}976 \\
    DiffusionDrive & 2{,}894 & 426 & 385 & 2{,}083 \\
    PRIX           & 2{,}677 & 374 & 337 & 1{,}966 \\
    \bottomrule
  \end{tabular*}
\end{minipage}
\end{table*}

\section{Trajectory-Only and Road-Only Alignment Variants}
\label{sec:supp-pairs}

In \Cref{fig:planners}, the detection teacher combines detected objects, navigation route, and future trajectory in its polar grid.
\Cref{tab:supp-pairs} compares this full detection teacher (\text{Seg.\ + Det.}) with two partial variants, both on top of semantic-segmentation guidance. In \text{Seg.\ + traj.\ pairs}, positive pairs are defined by the future trajectory alone. In \text{Seg.\ + road pairs}, they are defined by road geometry alone. In both variants, the alignment gradient does not reach the image encoder.
Over the baseline, trajectory pairs gain $+1.0$ on LTF, and road pairs gain $+0.9$ on MeanFuser and DrivoR.

\section{More Navhard Scenes}
\label{sec:supp-bev}

\Cref{fig:bev} shows scenes in which our module consistently helps.
\Cref{fig:supp-bev-drivor,fig:supp-bev-ltf,fig:supp-bev-dd} show uncurated scenes.
For each base model, we draw the same two runs as in \cref{fig:bev}: the baseline and our module (Seg.\ + Det.\ on
DrivoR, Seg.\ on LTF and DiffusionDrive), on the 5{,}912 navhard scenes.
A scene enters the pool if one of the two runs scores below 50 EPDMS (scenes that both handle are too easy to tell
them apart) and the two differ by more than 5 EPDMS (scenes that both fail alike show no difference).
Wins and losses of our module enter the pool alike.
We then shuffle the pool with a fixed seed and keep the first twelve scenes from distinct drives, since many navhard
scenes are perturbations of the same drive.

\Cref{tab:supp-bev-pool} shows how the two models compare over all scenes, not only the sampled ones.
The score of a single run on a single scene varies from seed to seed, so for each scene we average EPDMS over the five
seeds of each model and compare the two averages.
We count only the scenes in which the baseline or our module scores below 50 EPDMS, since both score at least 50 on the
remaining scenes (for DrivoR, 2{,}081 of the 5{,}912 scenes are counted).
On most counted scenes, the two models score within 5 EPDMS of each other (1{,}547 of the 2{,}081 on DrivoR).
On the rest, our module is higher by more than 5 EPDMS in more scenes than it is lower, on every base model
(\eg, 306 \vs 228 on DrivoR and 518 \vs 393 on LTF).
The advantage is therefore consistent across base models, though modest.
Rather than eliminating most failures, our module improves the outcome on a subset of scenes.
This is also why single scenes, such as those in the figures below, rarely show a large difference between the two
models.

Scores in the figures correspond to the displayed rollouts.
Because DiffusionDrive sampling is stochastic, these scores can differ slightly from the evaluation run.

\newlength{\suppw}
\setlength{\suppw}{\dimexpr(\textwidth-8pt)/4\relax}
\newcommand{\suppimg}[1]{\includegraphics[width=\suppw]{figs/bev/#1_cam.jpg} & \includegraphics[width=\suppw]{figs/bev/#1_bev.png}}
\newcommand{\supplab}[2]{\multicolumn{2}{c}{\scriptsize{\color{bevRef}Baseline: #1}\quad{\color{bevOurs}+ Ours: #2}}}

\begin{figure*}[p]
  \fycolor
  \centering
  \setlength{\tabcolsep}{0pt}
  \renewcommand{\arraystretch}{0.6}
  \begin{tabular}{@{}c@{\hspace{1pt}}c@{\hspace{6pt}}c@{\hspace{1pt}}c@{}}
    \suppimg{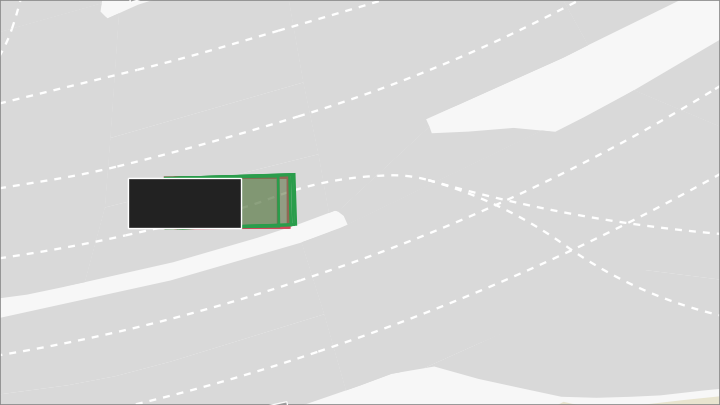} & \suppimg{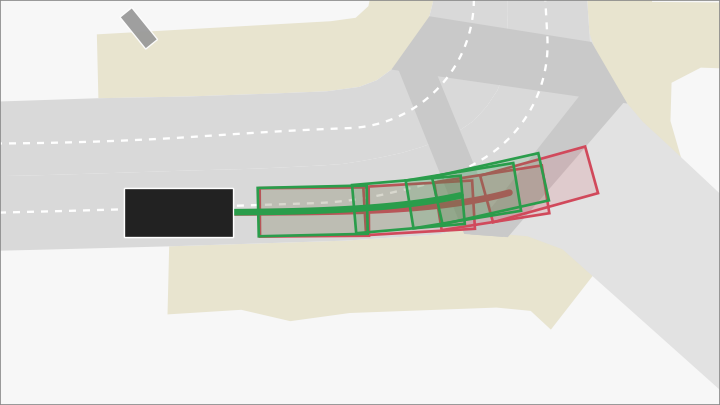} \\
    \supplab{EPDMS 0 (off-road)}{EPDMS 86} & \supplab{EPDMS 0 (off-road)}{EPDMS 66} \\[2pt]
    \suppimg{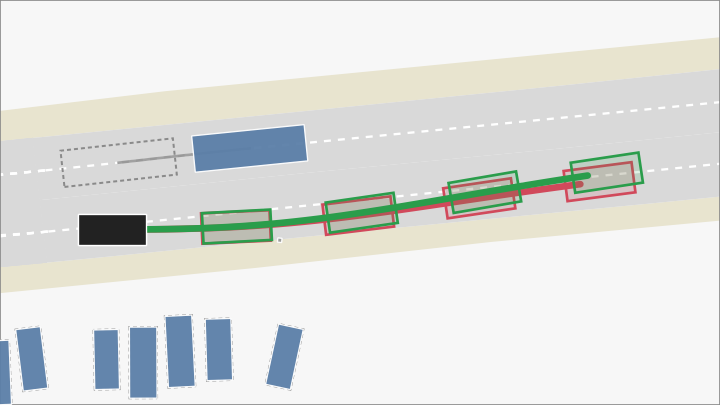} & \suppimg{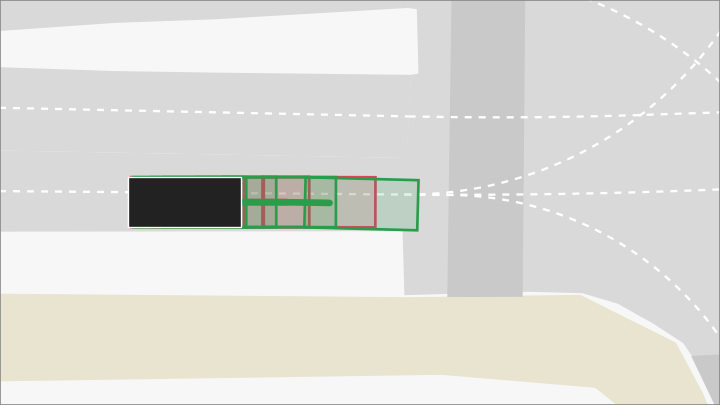} \\
    \supplab{EPDMS 0 (off-road)}{EPDMS 23} & \supplab{EPDMS 100}{EPDMS 0 (red light)} \\[2pt]
    \suppimg{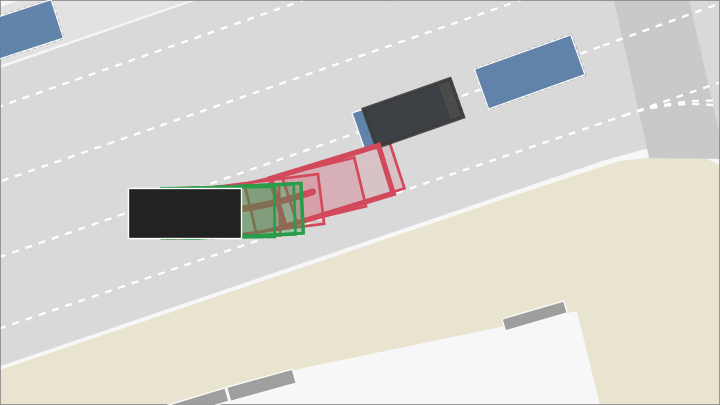} & \suppimg{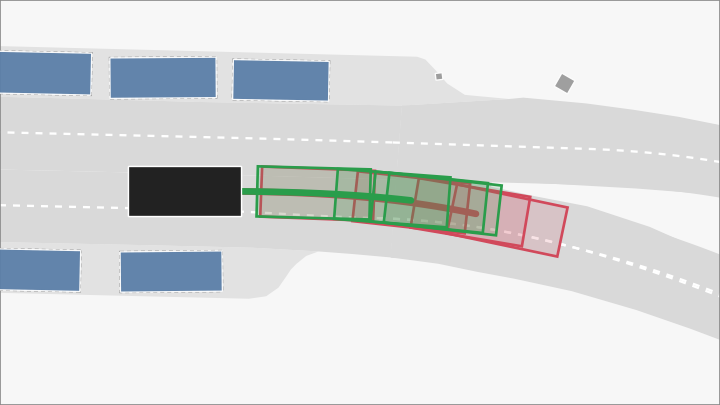} \\
    \supplab{EPDMS 0 (collision)}{EPDMS 68} & \supplab{EPDMS 79}{EPDMS 0 (off-road)} \\[2pt]
    \suppimg{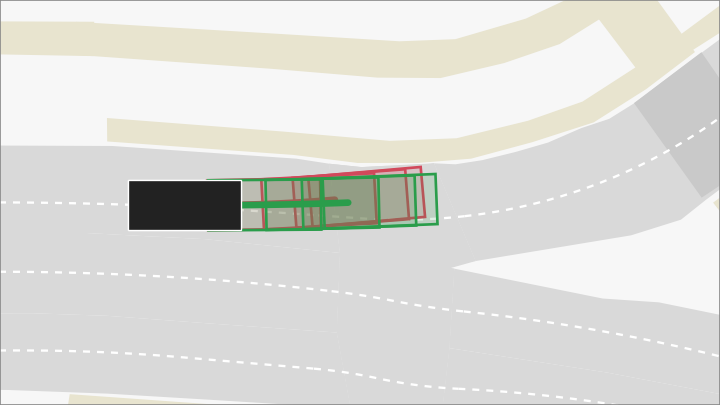} & \suppimg{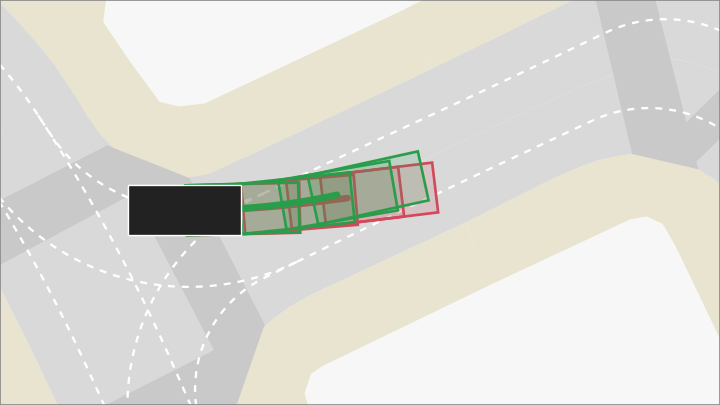} \\
    \supplab{EPDMS 40}{EPDMS 41} & \supplab{EPDMS 25}{EPDMS 25} \\[2pt]
    \suppimg{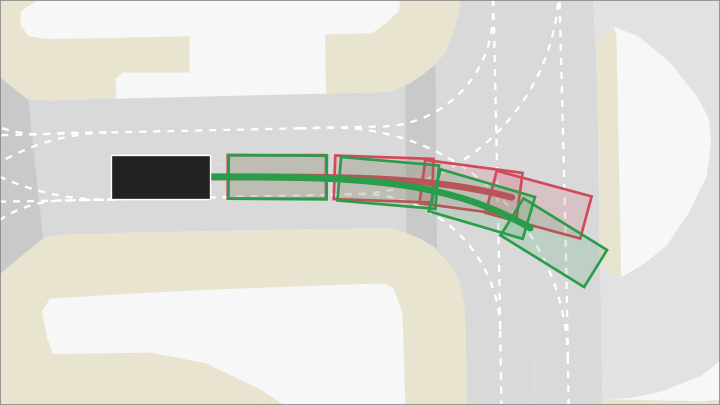} & \suppimg{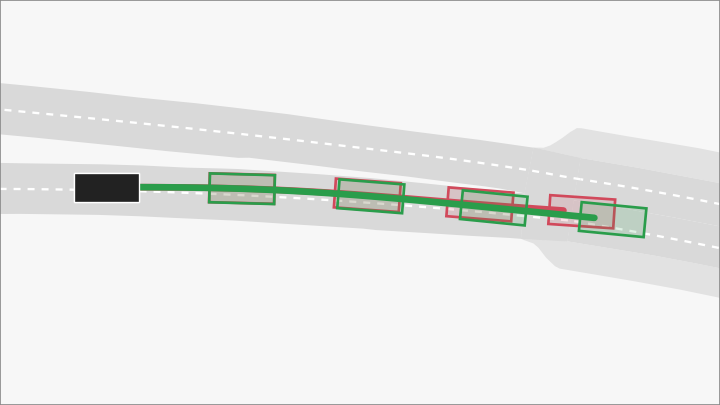} \\
    \supplab{EPDMS 96}{EPDMS 0 (off-road)} & \supplab{EPDMS 0 (off-road)}{EPDMS 97} \\[2pt]
    \suppimg{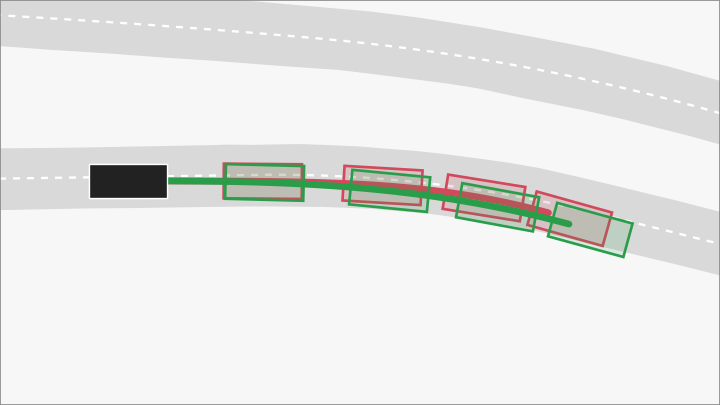} & \suppimg{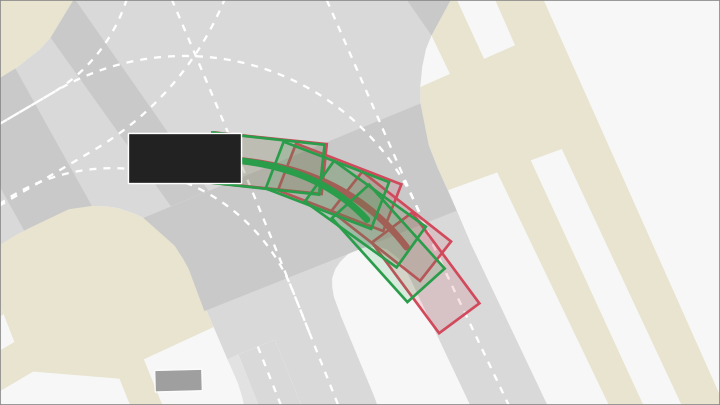} \\
    \supplab{EPDMS 90}{EPDMS 0 (off-road)} & \supplab{EPDMS 100}{EPDMS 0 (off-road)} \\[2pt]
  \end{tabular}
  \caption{DrivoR on 12 random navhard scenes (\cref{sec:supp-bev}), drawn as in \cref{fig:bev}:
  baseline ({\color{bevRef}red}) and with our module ({\color{bevOurs}green}; Seg.\ + Det.). Ours scores higher in 5 and lower in
  5 of them; 2 are within 5 EPDMS in the re-run.}
  \label{fig:supp-bev-drivor}
\end{figure*}

\begin{figure*}[p]
  \fycolor
  \centering
  \setlength{\tabcolsep}{0pt}
  \renewcommand{\arraystretch}{0.6}
  \begin{tabular}{@{}c@{\hspace{1pt}}c@{\hspace{6pt}}c@{\hspace{1pt}}c@{}}
    \suppimg{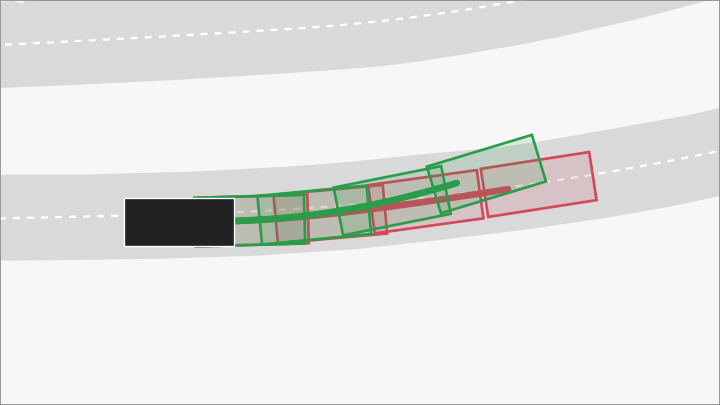} & \suppimg{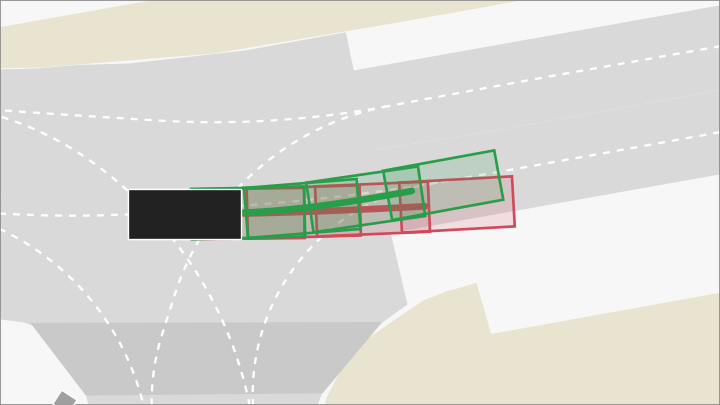} \\
    \supplab{EPDMS 89}{EPDMS 0 (off-road)} & \supplab{EPDMS 0 (off-road)}{EPDMS 83} \\[2pt]
    \suppimg{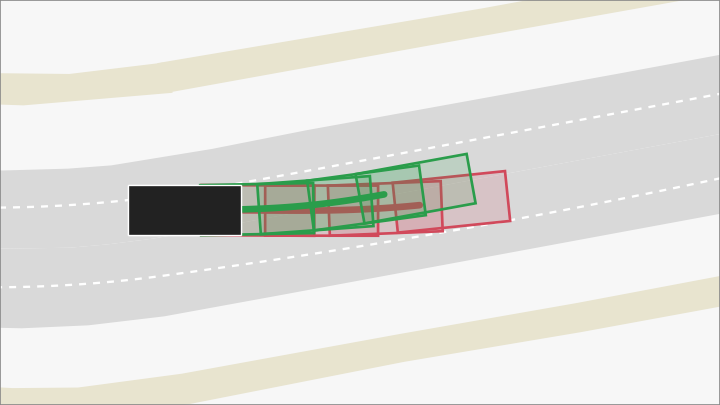} & \suppimg{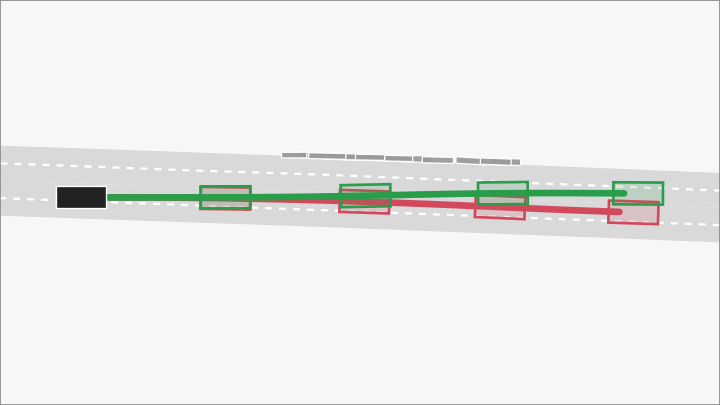} \\
    \supplab{EPDMS 34}{EPDMS 66} & \supplab{EPDMS 86}{EPDMS 0 (wrong direction)} \\[2pt]
    \suppimg{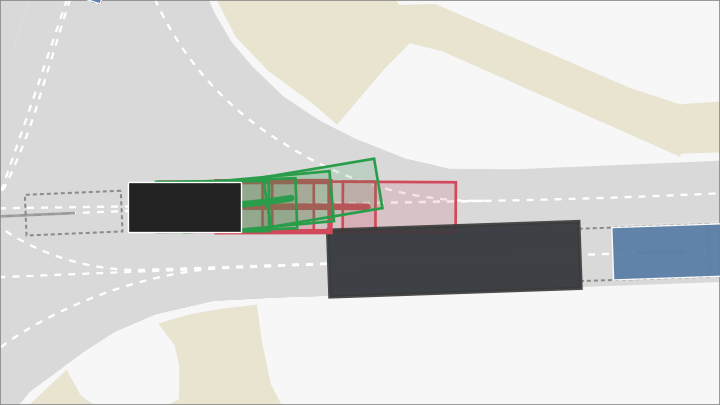} & \suppimg{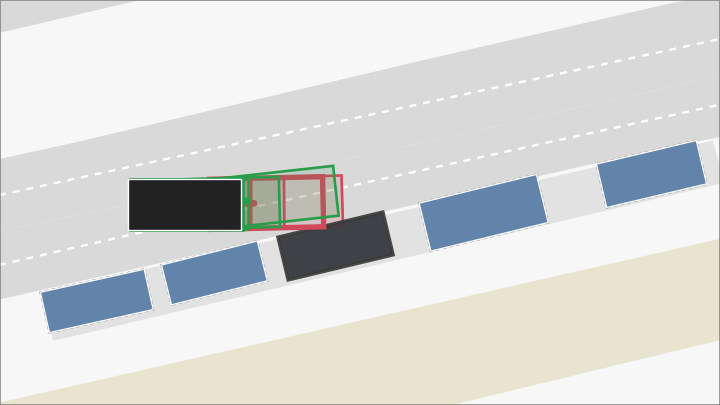} \\
    \supplab{EPDMS 0 (collision)}{EPDMS 100} & \supplab{EPDMS 0 (collision, off-road)}{EPDMS 29} \\[2pt]
    \suppimg{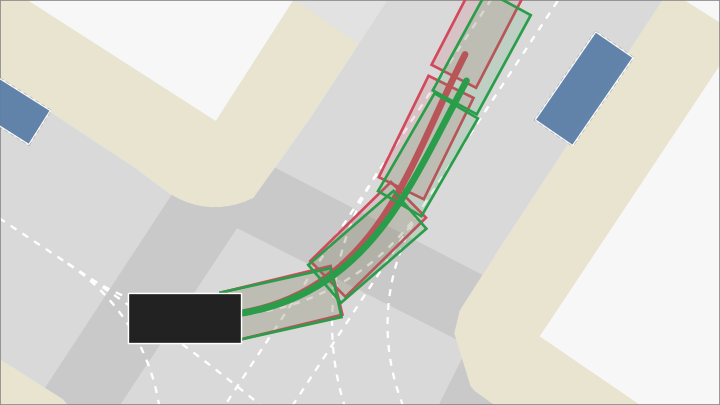} & \suppimg{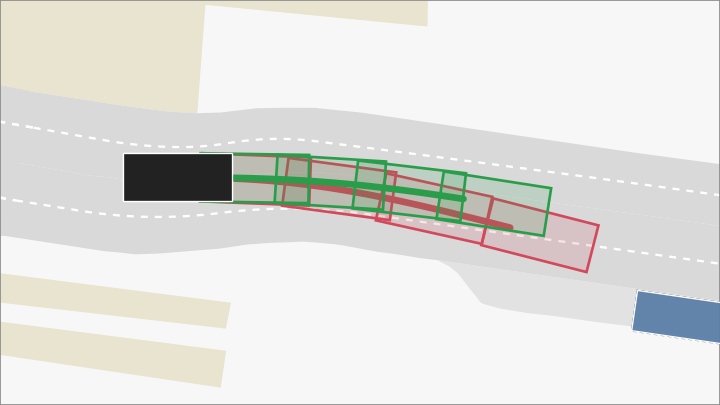} \\
    \supplab{EPDMS 0 (wrong direction)}{EPDMS 43} & \supplab{EPDMS 79}{EPDMS 37} \\[2pt]
    \suppimg{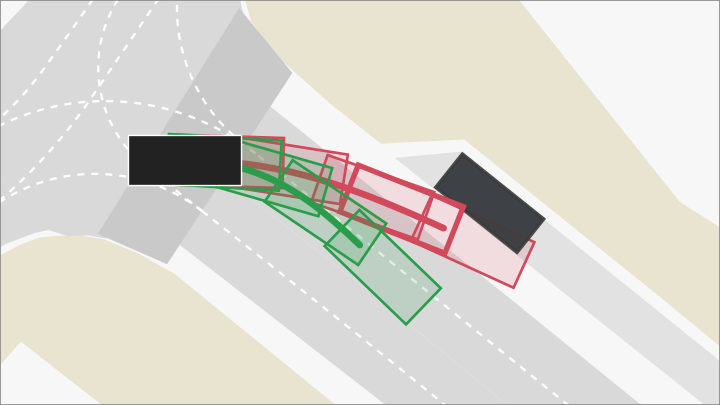} & \suppimg{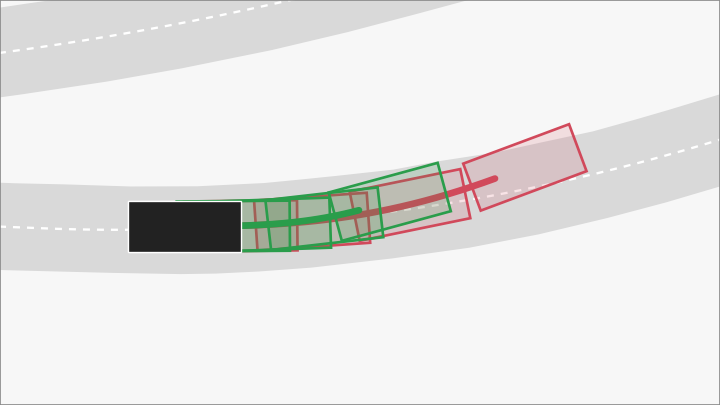} \\
    \supplab{EPDMS 0 (collision, off-road)}{EPDMS 43} & \supplab{EPDMS 0 (off-road)}{EPDMS 81} \\[2pt]
    \suppimg{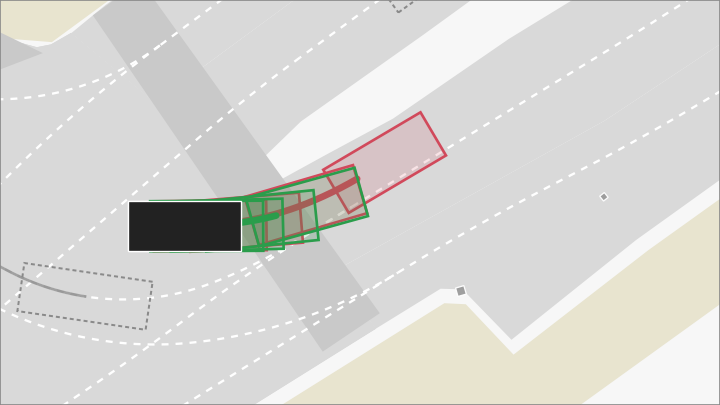} & \suppimg{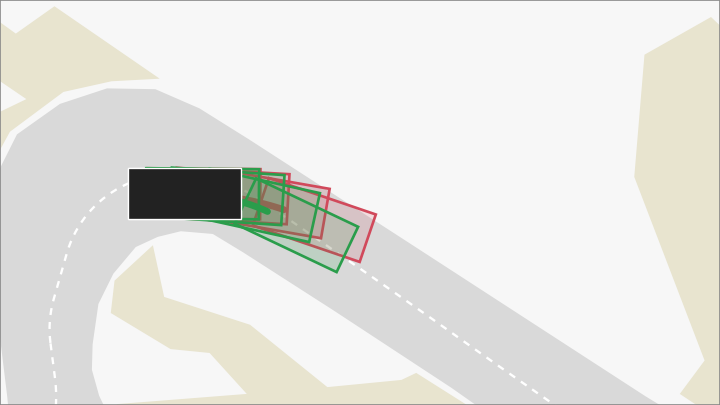} \\
    \supplab{EPDMS 44}{EPDMS 79} & \supplab{EPDMS 0 (off-road)}{EPDMS 72} \\[2pt]
  \end{tabular}
  \caption{LTF on 12 random navhard scenes (\cref{sec:supp-bev}), drawn as in \cref{fig:bev}:
  baseline ({\color{bevRef}red}) and with our module ({\color{bevOurs}green}; Seg.). Ours scores higher in 9 and lower in
  3 of them.}
  \label{fig:supp-bev-ltf}
\end{figure*}

\begin{figure*}[p]
  \fycolor
  \centering
  \setlength{\tabcolsep}{0pt}
  \renewcommand{\arraystretch}{0.6}
  \begin{tabular}{@{}c@{\hspace{1pt}}c@{\hspace{6pt}}c@{\hspace{1pt}}c@{}}
    \suppimg{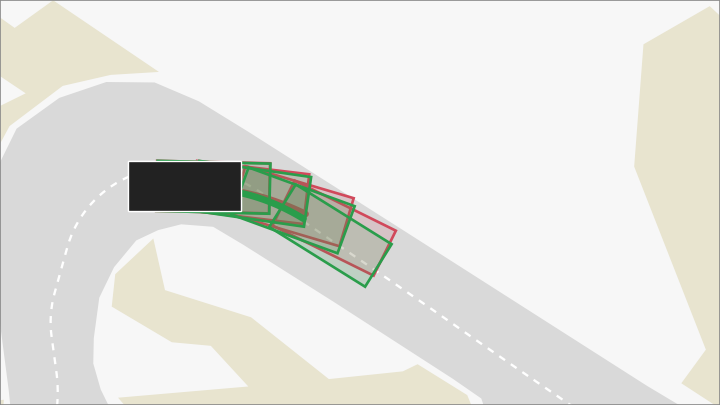} & \suppimg{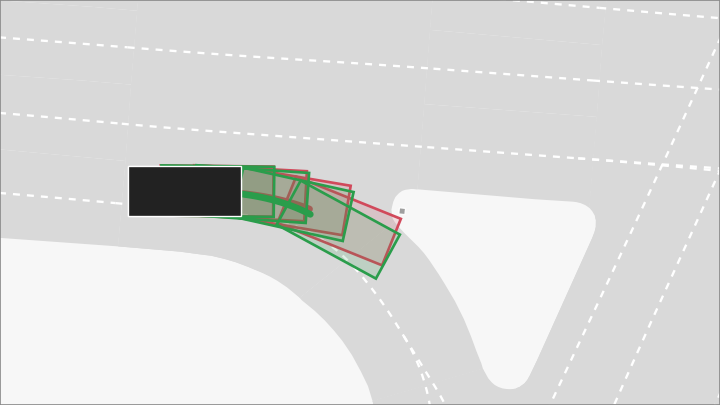} \\
    \supplab{EPDMS 0 (off-road)}{EPDMS 91} & \supplab{EPDMS 0 (off-road)}{EPDMS 91} \\[2pt]
    \suppimg{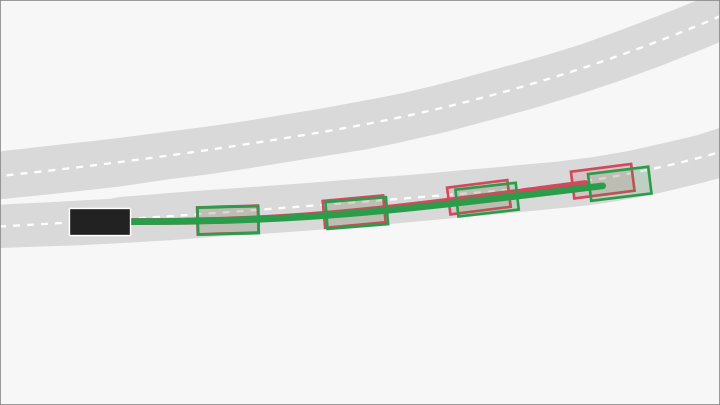} & \suppimg{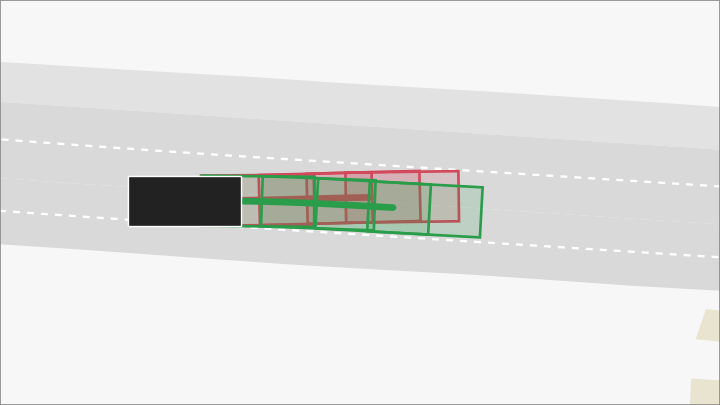} \\
    \supplab{EPDMS 85}{EPDMS 0 (off-road)} & \supplab{EPDMS 32}{EPDMS 66} \\[2pt]
    \suppimg{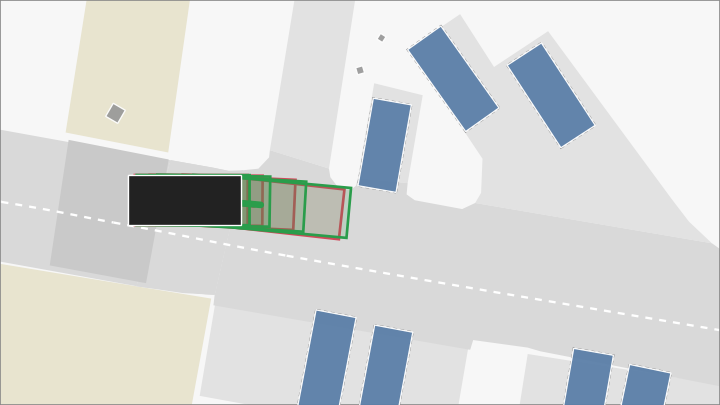} & \suppimg{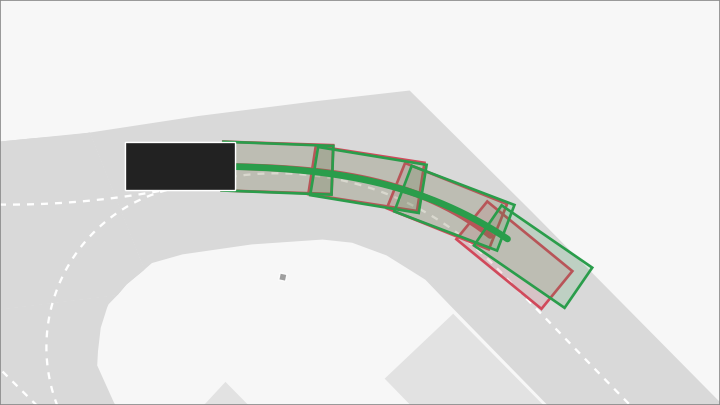} \\
    \supplab{EPDMS 79}{EPDMS 0 (off-road)} & \supplab{EPDMS 84}{EPDMS 0 (off-road)} \\[2pt]
    \suppimg{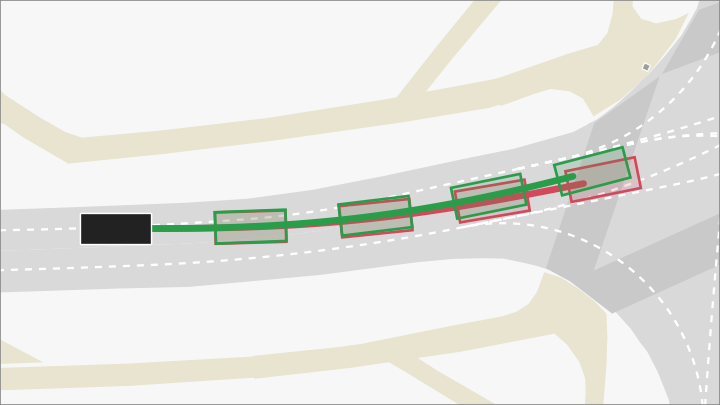} & \suppimg{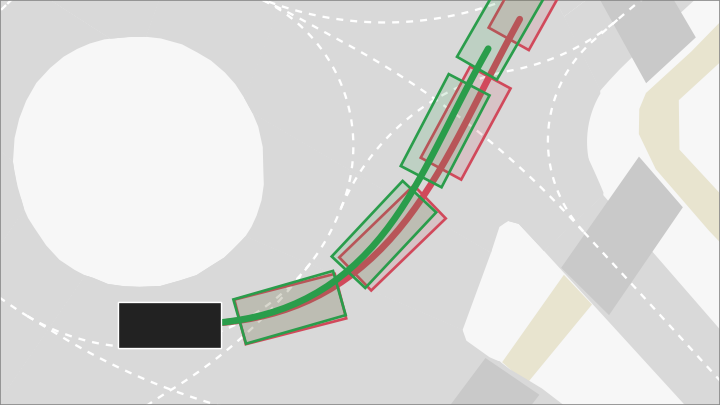} \\
    \supplab{EPDMS 0 (wrong direction)}{EPDMS 76} & \supplab{EPDMS 0 (off-road)}{EPDMS 76} \\[2pt]
    \suppimg{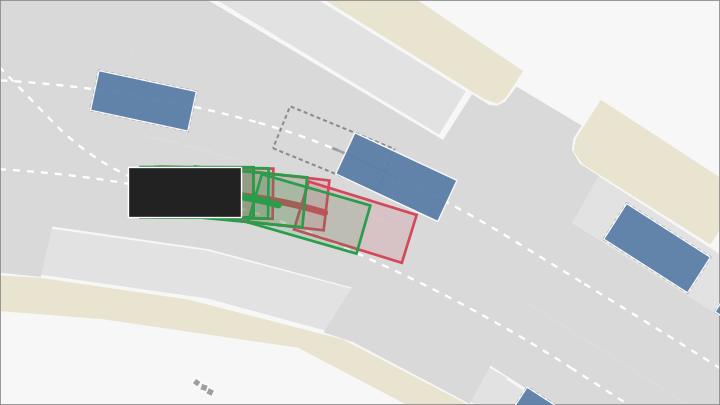} & \suppimg{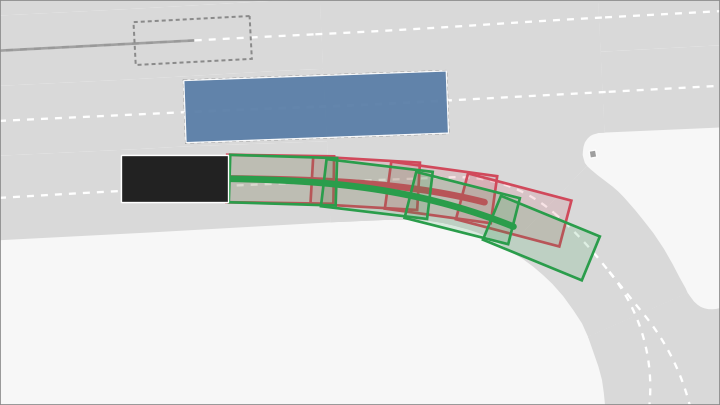} \\
    \supplab{EPDMS 51}{EPDMS 45} & \supplab{EPDMS 83}{EPDMS 0 (off-road)} \\[2pt]
    \suppimg{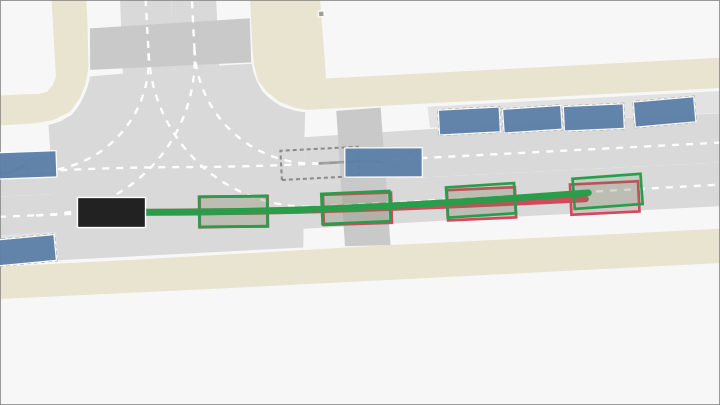} & \suppimg{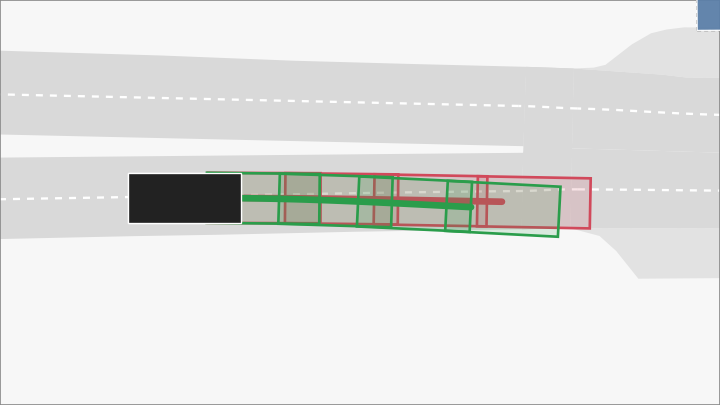} \\
    \supplab{EPDMS 0 (off-road)}{EPDMS 94} & \supplab{EPDMS 87}{EPDMS 0 (off-road)} \\[2pt]
  \end{tabular}
  \caption{DiffusionDrive on 12 random navhard scenes (\cref{sec:supp-bev}), drawn as in \cref{fig:bev}:
  baseline ({\color{bevRef}red}) and with our module ({\color{bevOurs}green}; Seg.). Ours scores higher in 6 and lower in
  6 of them.}
  \label{fig:supp-bev-dd}
\end{figure*}

\end{document}